\documentclass{article}

    \usepackage[final]{neurips_2024}

\usepackage[utf8]{inputenc} 
\usepackage[T1]{fontenc}    
\usepackage{hyperref}       
\hypersetup{
	colorlinks=true,       
	linkcolor=blue,        
	citecolor=blue,        
	filecolor=magenta,     
	urlcolor=blue         
}
\usepackage{url}            
\usepackage{booktabs}       
\usepackage{amsfonts}       
\usepackage{nicefrac}       
\usepackage{microtype}      
\usepackage{xcolor}         
\usepackage{xspace}  
\usepackage{graphicx}
\usepackage{amsmath}
\usepackage{amssymb}
\usepackage{mathtools}
\usepackage{amsthm}
\usepackage{listings}
\usepackage{natbib}
\usepackage{enumitem}
\usepackage{silence}
\usepackage[capitalize,noabbrev]{cleveref}
\usepackage[textsize=tiny]{todonotes}
\usepackage{indentfirst} 
\usepackage{microtype}
\usepackage{graphicx}
\usepackage{algorithm}
\usepackage{subfigure}
\usepackage{wrapfig}
\usepackage{booktabs} 
\usepackage{hyperref}
\usepackage{multirow}
\usepackage{worldflags}
\usepackage{makecell}
\usepackage{subcaption}
\usepackage{etoc}

\newcommand{\prompt}[1]{{\small \ttfamily #1}\xspace}
\newcommand{\llms}{LLMs\xspace}

\newcommand{\usflag}{\worldflag[width=2mm,framewidth=0mm]{US}}

\newcommand{\arflag}{\worldflag[width=2mm,framewidth=0mm]{AR}}
\newcommand{\esflag}{\worldflag[width=2mm,framewidth=0mm]{ES}}

\newcommand{\lily}{\usflag Lily\xspace}
\newcommand{\abdul}{\arflag Abdul\xspace}
\newcommand{\javier}{\esflag Javier\xspace}
\newcommand{\method}{CulturePark\xspace}

\title{CulturePark: Boosting Cross-cultural Understanding in Large Language Models}

\author{%
  Cheng Li\thanks{Work done during Cheng's internship at MSRA.} \\
  Institute of Software, CAS\\
  \texttt{chenglicat0228@gmail.com} \\
  \And
  Damien Teney \\
  Idiap Research Institute \\
  \texttt{contact@damienteney.info} \\
  \And
  Linyi Yang \\
  Westlake University \\
  \texttt{yanglinyi@westlake.edu.cn} \\
  \And
  Qingsong Wen \\
  Squirrel AI \\
  \texttt{qingsongedu@gmail.com} \\
  \And
  Xing Xie \\
  Microsoft Research \\
  \texttt{xing.xie@microsoft.com} \\
  \And
  Jindong Wang\thanks{Corresponding author. Work done at MSRA.} \\
  William \& Mary \\
  \texttt{jwang80@wm.edu} \\
}

\begin{document}

\etocdepthtag.toc{mtchapter}
\etocsettagdepth{mtchapter}{none}
\etocsettagdepth{mtappendix}{none}

\maketitle

\begin{abstract}

Cultural bias is pervasive in many large language models (\llms), largely due to the deficiency of data representative of different cultures.
Typically, cultural datasets and benchmarks are constructed either by extracting subsets of existing datasets or by aggregating from platforms such as Wikipedia and social media.
However, these approaches are highly dependent on real-world data and human annotations, making them costly and difficult to scale.
Inspired by cognitive theories on social communication, this paper introduces \textit{\method}, an LLM-powered multi-agent communication framework for cultural data collection.
\method simulates cross-cultural human communication with LLM-based agents playing roles in different cultures.
It generates high-quality cross-cultural dialogues encapsulating human beliefs, norms, and customs.
Using \method, we generated 41,000 cultural samples to fine-tune eight culture-specific \llms.
We evaluated these models across three downstream tasks: content moderation, cultural alignment, and cultural education.
Results show that for content moderation, our GPT-3.5-based models either match or outperform GPT-4 on $41$ datasets. Regarding cultural alignment, our models surpass GPT-4 on Hofstede's VSM 13 framework~\citep{vsm13} .
Furthermore, for cultural education of human participants, our models demonstrate superior outcomes in both learning efficacy and user experience compared to GPT-4. \method proves an important step in addressing cultural bias and advancing the democratization of AI, highlighting the critical role of culturally inclusive data in model training. Code is released at \url{https://github.com/Scarelette/CulturePark}.

\end{abstract}

\section{Introduction}

Culture is an important part of human society, composed of human beliefs, norms, customs, etc.~\citep{spencer2012culture}.
As large language models (\llms) play a vital role in daily communication~\citep{yang2024social}, recommendation systems~\citep{li2023prompt,fan2023recommender}, and education~\citep{shaikh2023rehearsal}, it is imperative for \llms to perceive and reflect different cultures.
However, current state-of-the-art \llms have been reported to be biased towards mainstream culture while ignoring others, resulting in a cultural bias problem~\citep{liu2023multilingual,cao2023assessing,masoud2023cultural,naous2023having,wang2023not,johnsonghost}. This leads to stereotypical impressions of different cultures, which can even exacerbate social conflicts~\citep{ryan2024unintended}.
The main reason behind cultural bias is that the training corpus of \llms is dominated by English data that express the cultural values and opinions of western people.
Much less can be learned about other cultures simply because there is less data available, i.e., a low-resource situation.

Existing approaches to solve the cultural bias problem in \llms include prompt engineering~\citep{kovavc2023large,wang2023not,rao2023ethical} and pre-training in non-English languages~\citep{pires2023sabia,chan2023harmonizing,nguyen2023seallms,pipatanakul2023typhoon,abbasi2023persianllama,lin2023taiwan}.
Prompt engineering consists of tuning prompts for different cultural tasks, but the benefits do not hold reliably across various downstream tasks.
Pre-training in various languages is promising, but the data collection and the pre-training itself are both very costly.
More importantly, cultural differences are embodied in many aspects such as opinions, customs, norms, and languages.
A model serving all cultures may face cultural conflict and misalignment problems \citep{liu2023multilingual,cao2023assessing,masoud2023cultural}.
Thus, it is necessary to fine-tune culture-specific models that target specific cultures.
Recently, \citet{li2024culturellm} proposed CultureLLM, which augments the fine-tuning data of \llms via semantic data augmentation to train culture-specific \llms.
However, the generated data lack diversity because it is implemented by generating semantically equivalent sentences of seed examples. 


\begin{figure}[t!]
    \centering
    \includegraphics[width=\textwidth]{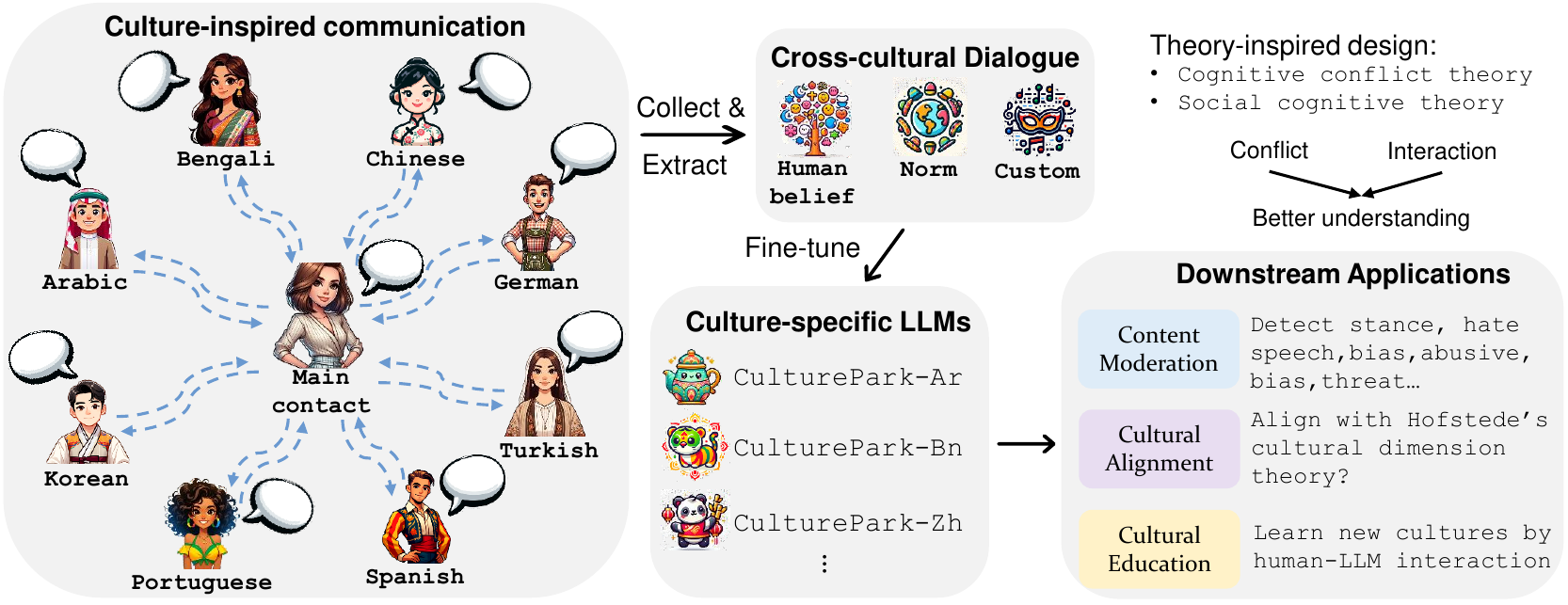}
    \caption{\method is an LLM-based multi-agent communication platform for cultural data collection. Leveraging \method, we can collect a cross-cultural dialogue dataset, which can then be used for fine-tuning culturally specific \llms to be applied to different downstream tasks: content moderation, cultural alignment, and cultural education.}
    \label{fig-overview}
\end{figure}

In this paper, we present \textbf{\method}, an LLM-powered multi-agent framework to simulate cross-cultural communication of humans. 
As shown in \Cref{fig-overview}, \method serves as an effective data collection platform to generate diverse and high-quality cultural datasets via multi-agent communication.
\method consists of a main contact (an English-speaking agent, 
\lily) who is in charge of the multi-turn dialogue and several cultural delegates (e.g., \abdul) who interact with the main contact and create cognitive conflicts.\footnote{We choose the English agent as the main contact since \llms do the best role-playing using English.}
After an initial problem is provided as input to the framework, the agents discuss the problem and express their opinions. 
Their different cultural backgrounds and genders boost diverse opinions and encourage one another to think more deeply. 
Original questions and ground truth can be augmented by creating novel questions and more comprehensive answers.
The interactions eventually generate a cross-cultural dialogue dataset that contains deep and comprehensive thinking and informative knowledge of different cultures. 
Detailed statistics are shown in \Cref{tb-observation}.
We then refine the original dataset to factually verify and increase its diversity, which is used to fine-tune culturally specific \llms for downstream tasks, as shown in \Cref{fig-pipeline-all}.

From the perspective of cognitive social science, our framework is inspired by Cognitive Conflict Theory (CCT)~\citep{limon2001cognitive,cosier1977cognitive} and Social Cognitive Theory (SCT)~\citep{fiske1991social} to foster a collaborative and communicative environment for mutual understanding of cultures.
Specifically, \method allows agents to encounter \emph{cognitive conflicts}, which triggers deeper thinking on certain topics according to CCT~\citep{limon2001cognitive,cosier1977cognitive}.
At the same time, a deeper understanding of cultures can be prompted through \emph{interaction and communication} with other agents, as suggested by SCT~\citep{fiske1991social}.
In favor of these theories, we found that \method triggers \llms' cross-cultural understanding ability, boosts novel opinions by allowing agents to think deeper, and benefits data augmentation by creating more comprehensive answers to the questions.
\method has the potential to facilitate the data collection of culture-related tasks, cultural value alignment, and improving AI democracy.





In summary, the contributions of this paper are three-fold:

\begin{enumerate}[leftmargin=2em]
\setlength\itemsep{0em}
    \item We introduce \method, a cost-efficient multi-agent framework to boost the cross-cultural understanding in \llms. Our platform creates cognitive conflicts and interactions between different cultural agents. More importantly, the platform uncovers several interesting findings, such as communication enables the cross-cultural understanding ability of \llms, boosts novel opinions, and benefits data augmentation.
    \item Leveraging \method, we generate and augment novel questions and more comprehensive answers, leading to $41K$ cultural samples in total. Those data contain rich and diverse information on different aspects of culture, such as norms, opinions, and backgrounds. We then fine-tune cultural specific \llms for different cultures.
    \item We evaluate \method in three key experiments: 1) The fine-tuned \llms outperforms GPT-4 in $5$ cultures on $26$ content moderation tasks and approach GPT-4 on other tasks; 2) Our fine-tuned \llms achieve better performance on cultural alignment experiments via Hofstede's cultural dimensions theory \citep{lamport94}; and 3) Human participants can perform more effective culture learning in situated learning experiments and show better satisfaction compared to GPT-4.
\end{enumerate}

\section{Related work}


\subsection{Cultural bias in \llms}


A body of research has explored cultural biases in \llms.  \citet{johnson2022ghost} examined conflicts in model outputs and input values, using moral value pluralism to analyze the responses of GPT-3 against global demographics. Their results showed that conflicting values were more aligned with the dominant US values reported. \citet{naous2023having} highlighted a bias towards Western culture in models processing Arabic, exacerbated by English-aligned prompts, suggesting mitigation through cultural tokens. The Cultural Alignment Test (CAT), based on Hofstede’s framework \citep{lamport94}, evaluated cultural values in models such as ChatGPT and Bard across different cultures, revealing the highest cultural alignment for GPT-4 with US values~\citep{masoud2023cultural}. \citet{cao2023assessing} found that ChatGPT aligned well with American culture but struggled with other cultures, particularly under English prompts. Additionally, \citet{liu2023multilingual} reported that multilingual LLMs had limited abilities to reason with proverbs and exhibited a ``culture gap'' in handling translations, leading to the development of the MAPS dataset for assessing proverb comprehension in six languages.

\subsection{Cultural benchmarks and datasets}

Extensive research has focused on developing cultural benchmarks, which can be categorized into two types: collecting existing datasets and synthesizing new ones.
First, most of the work adopted existing datasets as sources of cultural data.
\citet{wang2023not} introduced a benchmark that uses cultural items to analyze cultural dominance, based on sources such as WVS~\citep{wvs} and PCT~\citep{mudde20162012}.
Later work includes Cultural Alignment Test \citep{masoud2023cultural}, NORMSAGE \citep{fung2022normsage}, WorldValueBench \citep{zhao2024worldvaluesbench}, and NORMAD \citep{rao2024normad} that sourced from different existing datasets.
Other types of data sources include CultureAtlas \citep{fung2024massively} and MAPS~\citep{liu2023multilingual} which collected data from Wikimedia; Candle~\citep{nguyen2023extracting} and CultureBank~\citep{shi2024culturebank} sourced their data from social media such as Tiktok and Reddit.
In contrast, there was an emerging trend to perform data augmentation for cultural \llms.
\citet{li2024culturellm} proposed semantic data augmentation to synthesize cultural data by enriching the semantic equivalence of the generated samples.

\method significantly differs from those that perform direct data collection from existing datasets; it also differs from CultureLLM \citep{li2024culturellm}  since \method leverages multi-agent communication for data generation, which is more natural and can generate more diverse datasets.

\subsection{Existing solutions to cultural bias}

There are primarily two types of approach to addressing the problem of cultural bias: prompt engineering and pre-training.
The work of \citep{kovavc2023large,wang2023not} viewed \llms as amalgamations of cultural perspectives, which can be oriented toward specific cultural perspectives through prompt engineering.
In contrast, Rao et al. \citep{rao2023ethical} integrated cultural values directly into the prompts. Although prompt engineering is cost-effective, its efficacy is questionable, particularly in low-resource cultures where \llms may lack relevant cultural knowledge due to underrepresentation in pre-training data.
An alternative strand of research focuses on pre-training and fine-tuning~\citep{pires2023sabia,chan2023harmonizing,nguyen2023seallms,pipatanakul2023typhoon,abbasi2023persianllama,lin2023taiwan}.
Those approaches developed culturally aware LLMs for various cultures by assembling large-scale pre-training datasets, followed by fine-tuning to enhance alignment. Despite achieving significant performance improvements, these methods were both costly and time consuming, making them impractical for a broader application across numerous cultures and countries. Furthermore, they still face challenges in low-resource cultures where acquiring pre-training data is difficult. For example, MaLA-500~\citep{lin2024mala} aimed to train a new LLM in Llama 2 to support $534$ languages, illustrating the resource-intensive nature of this approach. Unlike these approaches, \method provides a cost-effective solution to the cultural bias problem, including data augmentation and fine-tuning.

\section{\method}
\label{sec-method}

\subsection{Design}


\method is an LLM-powered\footnote{We use GPT-3.5-Turbo in this work due to its great performance, high efficiency, and manageable price. \method is a flexible platform that naturally supports other \llms such as GPT-4 and Llama models. More importantly, \method allows to use \textit{different} LLMs for the main contact and delegate (e.g., GPT-3.5 for main contact and Llama-2 for delegate), which makes it flexible to extend to other models and evaluate the cross-model understanding ability.} cross-cultural communication framework that generates data to support culture-related research such as building cultural-specific \llms and performing cultural alignment.
It is inspired by Cognitive Conflict Theory (CCT) and Social Cognition Theory (SCT) to design multi-turn communications for a deeper understanding of cultural topics.
CCT posits that cognitive conflicts can help individuals engage more in deeper thinking~\citep{limon2001cognitive,cosier1977cognitive}, and SST emphasizes that individuals can deepen their understanding of perspectives through explanation and debate~\citep{fiske1991social}.

\Cref{fig-overview} shows the overview of \method.
To enable English-based interaction, we design two types of cultural agents: the main contact and the cultural delegate.
Specifically, the main contact agent, \lily, is from English culture and is responsible for all conversations with delegates from different cultures such as \abdul from Arabic and \javier from Spanish culture.
The complete information of agents and culture is in \Cref{tb-agent-name}.
As shown in \Cref{fig-pipeline-1}, we input a system prompt to \llms which contains the background setting and initial question to initiate the conversation.
The initial question, such as ``\prompt{How do you think about one of my main goals in life has been to make my parents proud? Please provide your opinions and reasons}'', is obtained from WVS~\citep{wvs} and GAS~\citep{pew}, two popular cultural surveys whose examples are shown in \Cref{fig-seed-data}.
After that, the agents conduct cross-cultural conversations to generate data.
Currently, \method supports $8$ cultures and $2$ genders while more cultures can be easily added.
Those agents could conduct in-cultural or cross-cultural communication, while we rely on cross-cultural more since in-cultural communication will likely generate less diverse topics (e.g., \Cref{fig-case-in-culture}).
We discuss the quality of data from in-cultural and cross-cultural communication and the influence of gender in \Cref{sec-discussion}.

We designed improved prompting techniques to maintain high-quality conversations.
First, the cultural bias of the main contact and cultural delegate is reduced by designing \textit{self-calibration} prompts to calibrate their outputs. 
We use a seed datum that contains the attitude of the target culture to the input question to guide the dialogue.
All the following statements should conform to the answer in seed.
As shown in \Cref{fig-pipeline-1}, we introduce the opinion from \abdul's culture and ask \abdul and \lily to conform to their cultures. The effect of the self-calibration prompt is shown in \Cref{fig-case-calibration-1,fig-case-calibration-2}. Without self-calibration prompts, \abdul's opinions contradict with Arabic people.
Second, the redundancy of the output, i.e., \llms always generates similar dialogues after multi-turn communication.
We devise two communication styles: one is \textit{self-guidance} prompts which can direct the dialogue to generate more diverse and informative data, such as ``\prompt{Are there anything in your culture related to the problem talked before}?" and ``\prompt{Do you agree with her? Provide more reasons to support your idea}?", and the other is free chat that does not need humans to participate and motivate the inner creativity of \llms. \Cref{fig-case-control,fig-case-free} show cases of self-guidance prompting and free chat, respectively.

\subsection{Data refinement and fine-tuning}



The seed questions initiating the communication have two sources: World Values Survey (WVS)~\citep{wvs} and Global Attitudes surveys (GAS) from Pew Research Center~\citep{pew}. 
WVS is a global research project that explores people's beliefs and values worldwide, examining how these beliefs evolve over time.
Pew Research Center, a nonpartisan organization, provides data and research on public opinion, social issues, and demographic trends both in the U.S. and globally. Its Global Attitudes surveys cover a wide range of topics, including politics, media, technology, religion, race, and ethnicity.
In total, we select $4.1$k seed data and generate $41$k dialogues (each dialogue contains several sentences).
We show the details of the data numbers for different cultures in \Cref{tb-data-num}. 
We also performed a statistical analysis on the GPT-4-based dataset.
As summarized in \Cref{fig-dataset}, the dataset contains human belief (59.68\%), norm (29.54\%) and custom (10.78\%) involving data on $8$ different cultures. 
\Cref{fig-seed-data,fig-case-control} show some examples of the seed data and the generated dialogues.

\begin{figure}[t!]
  \centering
  \subfigure[Cross-cultural dialogue dataset]{
    \includegraphics[width=.477\textwidth]{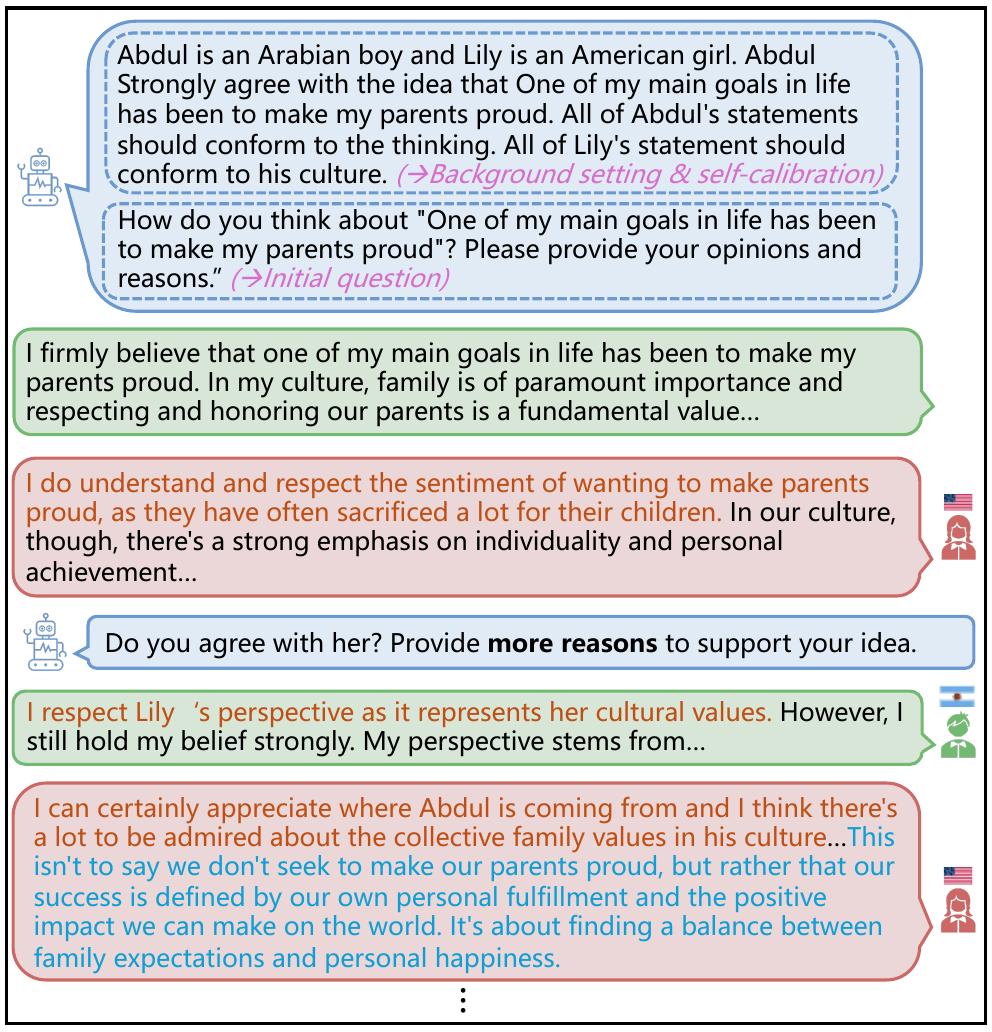}
    \label{fig-pipeline-1}
  }
  \subfigure[Data refinement for fine-tuning]{
    \includegraphics[width=.45\textwidth]{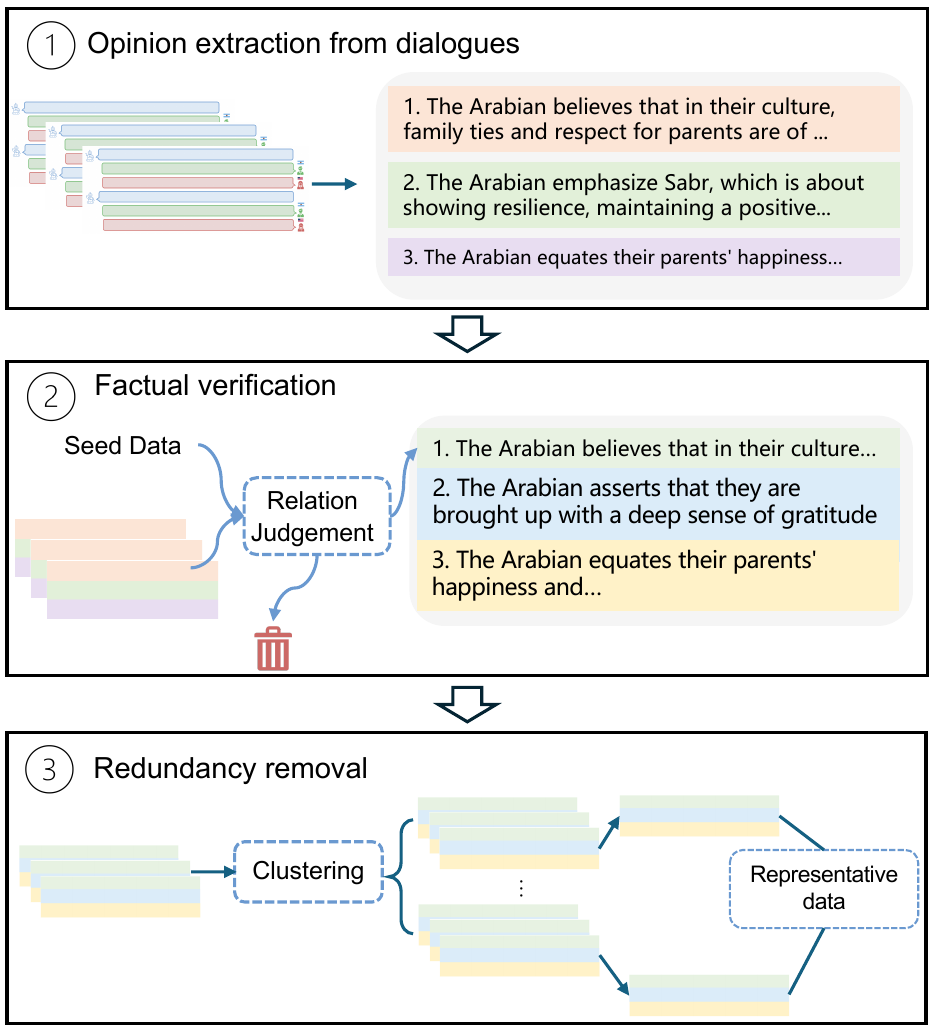}
    \label{fig-pipeline-2}
  }
  \caption{Cross-cultural dialogue and data refinement for fine-tuning \llms using \method.}
  \label{fig-pipeline-all}
\end{figure}

The generated dataset may not be directly used for fine-tuning since it could contain redundant and incorrect information that should be handled.
As shown in \Cref{fig-pipeline-2}, we design data refinement to refine the dataset.
First, the opinions on target culture are extracted from the dialogues generated via GPT-4, such as ``\prompt{The Arabian equates their parents' happiness and satisfaction to their own success}'' and ``\prompt{The Arabian emphasize Sabr, which is about showing resilience, maintaining a positive attitude and having faith during difficult times}''. 
Second, several extracted opinions could be irrelevant to the initial question or contradict with seed data, motivating us to perform verification to reserve only highly related opinions.
Furthermore, since the generated data could be semantically similar, we remove redundant samples to improve diversity.
To be specific, we obtain sentence embeddings via text-embedding-3-small~\citep{text-embedding-3-small} and cluster the embedding using K-means.
We reserve one sample for each cluster as representative data.
Eventually, we get the high-quality cultural data for different cultures.
The ablation of the refinement is in \Cref{tb-ablation-study}.

Algorithm~\ref{algo-aug} shows the pipeline for data refinement.
After refinement, there are $41$k samples (input-output pairs) left for fine-tuning, i.e., one sample for one dialogue.
Examples of the samples are provided in \Cref{fig-case-refine}.
Afterwards, we can fine-tune cultural-specific \llms using either open-source \llms or fine-tuning service.
In this paper, we mainly use OpenAI API to fine-tune GPT-3.5-Turbo due to its efficiency. Hyperparameters are shown in \Cref{tb-finetune}.
We further provide fine-tuning experiments on Llama2-70b in \Cref{sec-discuss-llama}.


\subsection{\method benefits cultural understanding and fine-tuning}

There are some interesting observations in communication among agents from different cultures.

\textbf{Communication triggers \llms' cross-cultural understanding ability.}
We observed that agents try to understand each other's opinions and the reasons sourced from their different cultural backgrounds.
For example, the blue sentences in \Cref{fig-pipeline-1} show cross-cultural understanding ability of \llms, such as ``\prompt{I do understand and respect the sentiment of wanting to make parents proud, as they have often sacrificed a lot for their children}''.
Leveraging GPT-4-Turbo, we analyzed the topics in the dataset such as human beliefs, norms, and customs, which can be further used as data collections for building culturally specific models.
\Cref{sec-append-dialogue} shows the details of the dialogue dataset, indicating that the generated topics are mostly about culture.
Then, we randomly sampled $750$ dialogues for each culture and evaluated the communication using the prompts in \Cref{sec-append-prompt}.
As summarized in \Cref{tb-observation}, on average, the ratio of statements that express cross-cultural understanding is $80.80$\%.
The analysis also verifies the effectiveness of \method in expanding topics and cross-cultural understanding.

\textbf{Cultural differences boost novel opinions.}
In cross-cultural communication, different opinions can inspire others to think deeper and more comprehensively, as suggested by CCT and SCT.
A case is the sentences in orange in \Cref{fig-pipeline-1}. \lily partially agrees with \abdul and gives an accurate and high-level summary of her pursuit: ``\prompt{a balance between family expectations and personal happiness}'' which is generated after a multi-turn energetic discussion with \abdul.
This also aligns well with CCT and SCT that emphasize the significance of communication among people having different cultural backgrounds.







\textbf{\method naturally assists cultural data augmentation by creating novel questions and comprehensive answers.}
On the one hand, agents in different cultures can generate new opinions towards certain topics, which intuitively diversifies the input questions.
On the other hand, the initial seed data only contain short answers such as ``\prompt{Strongly agree}'' with no further explanations.
Our platform allows deeper and more comprehensive communication of agents, thus generating more detailed responses such as ``\prompt{Strongly agree. I believe that pleasing parents and elders is a sign of respect and love}'' and ``\prompt{Strongly agree. I equate my parents' happiness and satisfaction to my own success}''.
Additionally, agents can extend the topics that conflict with their own opinions and provide more informative evidence to support their viewpoints.
This strategy helps to generate informative and diverse data continuously.
\Cref{sec-discuss-why} presents some detailed results on diversity gain, showing that the generated data has significantly larger diversity.

\section{Experiments}
\label{sec-exp}

\subsection{Evaluation on content moderation tasks}
\label{sec-exp-content}

\paragraph{Setup.}
Content moderation is crucial to maintaining the integrity and safety of online platforms in different cultures.
What is acceptable in one culture could be offensive or inappropriate in another. However, few methods focused on content moderation for different cultures.
For this experiment, we evaluated the effectiveness of our cultural-specific models for $8$ different cultures: Arabic, Bengali, Chinese, German, Korean, Portuguese, Spanish, and Turkish culture.
These cultures have their unique characters, involving a large number of people in the world.

We evaluate on $7$ content moderation tasks for $8$ different cultures to detect the following content: hate speech, offensive language, spam speech, abusive speech, bias speech, threat speech, and stance of speech in zero-shot evaluation, whose metric is average F1 score.
The details on the datasets can be found in \Cref{sec-append-dataset}.
In total, our test set contains $48,895$ samples. 
We compare our models with seven baselines: GPT-3.5-turbo~\citep{chatgpt}, GPT-4~\citep{openai2023gpt4}, Gemini-pro~\citep{gemini}, SeaLLM~\citep{nguyen2023seallms}, TaiwanLLM~\citep{lin2023taiwan}, Synatra-7B-v0.3-dpo~\citep{Synatra}, EEVE-Korean-10.8B-v1.0~\citep{EEVE}, CultureLLM~\citep{li2024culturellm}, and CultureBank~\citep{shi2024culturebank}.
CultureLLM is a series of culture-specific \llms using semantic data augmentation.
SeaLLM focuses on the Southeast Asian (SEA) culture, which is adopted for Chinese and Korean cultures.
TaiwanLLM focuses on traditional Chinese culture. Synatra-7B-v0.3-dpo and EEVE-Korean-10.8B-v1.0 targeted at Korean culture.
CultureBank collects data from social media and we compare Arabic and Korean culture by fine-tuning GPT-3.5-turbo on its dataset.

\begin{center}
\begin{minipage}{0.55\textwidth}
    \centering
    \includegraphics[width=\textwidth]{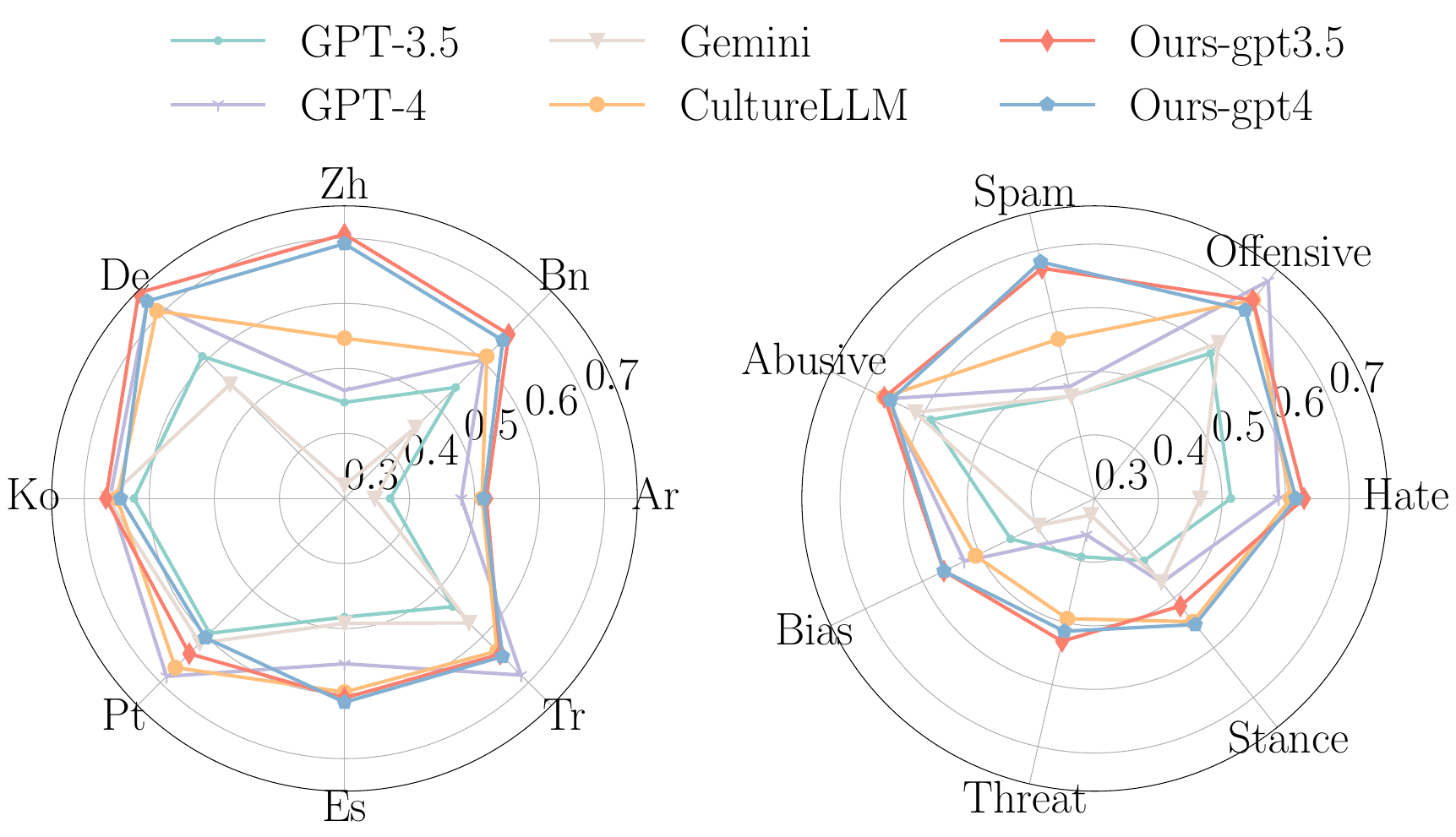} 
    \vspace{-.2in}
    \captionof{figure}{Results on content moderation.}
    \label{fig-content-moderation}
\end{minipage}
\hspace{.2in}
\begin{minipage}{0.35\textwidth}
    \centering
    \captionof{table}{Comparison with the latest cultural specific LLMs.}
    \label{tb-compare-llm}
    \resizebox{\textwidth}{!}{
    \begin{tabular}{llll}
        \toprule
        Chinese & Bias & Spam & Avg \\ \midrule
         SeaLLM & .237 & .357 & .297 \\ 
         Taiwan\_LLM & .446 & .341 & .394 \\ 
         Ours & \textbf{.530} & \textbf{.854} & \textbf{.692} \\ \midrule
        Arabic & Hate & Offensive & Avg \\ \midrule
         CultureBank & .540 & .642 & .591 \\ 
         Ours & \textbf{.558} & \textbf{.735} & \textbf{.602} \\ \midrule
        Korean & Abusive & Hate & Avg \\ \midrule
         Synatra-7B-v0.3-dpo & .390 & .465 & .428 \\ 
 EEVE-Korean-10.8B-v1.0 & .364 & .437 & .560 \\ 
         CultureBank & .635 & .522 & .579 \\
         Ours & \textbf{.647} & \textbf{.640} & \textbf{.643} \\ \bottomrule
    \end{tabular}
    }
\end{minipage}
\end{center}

\begin{wraptable}{r}{.45\textwidth}
\vspace{-.1in}
\caption{Results on ablation study of data generation and refinement.}
\label{tb-ablation-study}
\resizebox{.45\textwidth}{!}{
\begin{tabular}{l|cc|cc}
\toprule
\multicolumn{1}{c}{Model}                          & \multicolumn{1}{c}{Ar}     & \multicolumn{1}{c}{Bn}     & \multicolumn{1}{c}{Zh}     & \multicolumn{1}{c}{Pt}     \\ \midrule
\multicolumn{1}{c}{GPT-3.5-turbo}                   & \multicolumn{1}{c}{.370} & \multicolumn{1}{c}{.542} & \multicolumn{1}{c}{.448} & \multicolumn{1}{c}{.593} \\
\multicolumn{1}{c}{Generate}                  & \multicolumn{1}{c}{.451} & \multicolumn{1}{c}{.622} & \multicolumn{1}{c}{.636} & \multicolumn{1}{c}{.594} \\
\multicolumn{1}{c}{Generate+Verify}           & \multicolumn{1}{c}{.486} & \multicolumn{1}{c}{.635} & \multicolumn{1}{c}{.678} & \multicolumn{1}{c}{.604} \\
\multicolumn{1}{c}{Generate+Verify+Diversify} & \multicolumn{1}{c}{.514} & \multicolumn{1}{c}{.644} & \multicolumn{1}{c}{.692} & \multicolumn{1}{c}{.603} \\ \bottomrule
\end{tabular}
}
\end{wraptable}

\paragraph{Main Results.}
We analyzed the results from the culture and task sides in \Cref{fig-content-moderation}.
The most interesting observation is that our models outperformed GPT-4 on $5$ cultures and approached GPT-4 on the remaining $3$ cultures, although the data for fine-tuning are generated by GPT-3.5-turbo, which is much worse than GPT-4.
We also generated cultural data via GPT-4 and fine-tuned other $8$ cultural-specific models for comparison, denoted as ``Ours-gpt4'' in \Cref{fig-content-moderation}. The performance of those models is better than ``Ours'' (GPT-3.5-turbo version) but not so much. 
For other baselines, our models outperform them in most cases. \Cref{tb-compare-llm} shows that our models achieved better performance than those costly \llms which require pre-training and fine-tuning.

\paragraph{Ablation study.}
\Cref{tb-ablation-study} presents our ablation study on $4$ cultures, where
``Generate'' means just extracting opinions from the dialogue, ``Verify'' represents factually verifying the extracted opinions, and ``Diversify'' means removing redundant data.
The results show that each module of \method is effective, ensuring its interpretability.

\subsection{Evaluation on cultural alignment via Hofstede's Cultural Dimensions Theory}
\label{sec-exp-align}

\paragraph{Setup.}
Hofstede's cultural dimension theory is a framework for understanding cultural differences across countries based on data collected from various countries.
We asked \llms to answer the $24$ questions in VSM 13 to evaluate cultural alignment.
Specifically, we used a system prompt ``\prompt{You are a {culture} chatbot that knows {culture} very well}'' to induce \llms' cultural understanding ability.
We selected proper $C$\footnote{$C$ is constants that can be used to adjust scores to fit a range between $0$ and $100$ or anchor new data to Hofstede's old dataset~\citep{lamport94}.} and anchor \llms' answer to Hofstede's old dataset. We compute the gaps between \llms' answer and Hofstede's data from six cultural dimensions using the Euclidean distance.
Details on the survey and the distance are in \Cref{sec-append-alignment}.

\begin{wrapfigure}{r}{0.45\textwidth}
\vspace{-.2in}
  \begin{center}
    \includegraphics[width=.45\textwidth]{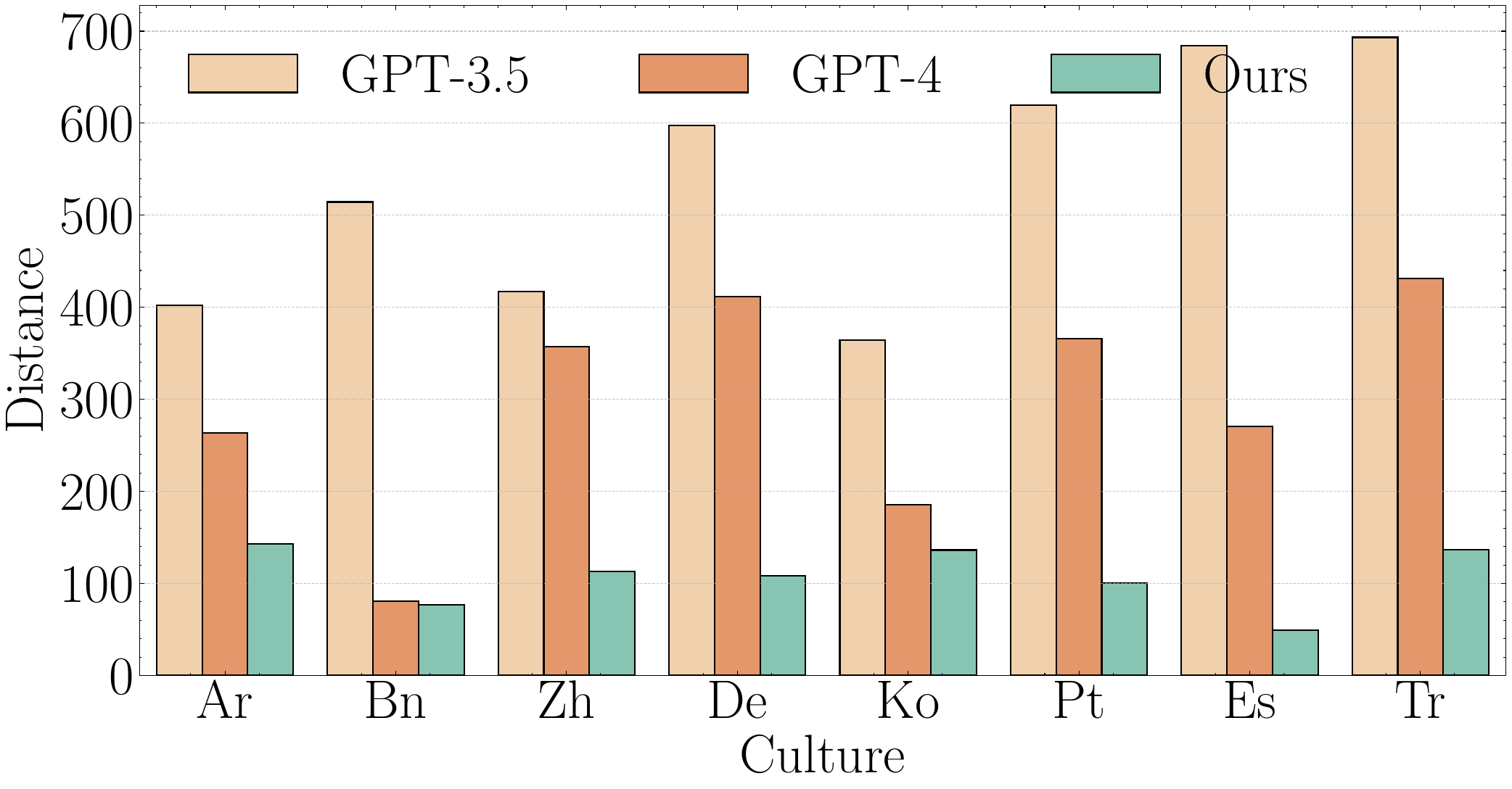}
  \end{center}
  \vspace{-.1in}
  \caption{Results on culture alignment via Hofstede's Cultural Dimensions Theory.}
  \label{fig-hofstede-res}
  \vspace{-.2in}
\end{wrapfigure}

\paragraph{Results.}
We compared our models (powered by GPT-3.5-Turbo) with GPT-3.5-turbo and GPT-4.
As shown in \Cref{fig-hofstede-res}, our models outperform them by a large margin, indicating their excellent cultural alignment and cultural understanding abilities.
Note that VSM is widely adopted as datasets for value and culture alignment, the results imply that our approach for data collection is effective, thus it could be further used for value alignment research.

\subsection{Evaluation in situated learning for cultural education}

Situated learning suggests that learning is best understood and facilitated when it occurs within the context \citep{anderson1996situated,lave1991situated}.
Motivated by situated learning, we leveraged \method for cultural education where our fine-tuned models serve as foreigners to talk to people about cultural problems, which can create a situation for cross-cultural communication and learning cultural-specific knowledge. 
For example, a person who wants to learn about Arabic culture can communicate with our Arabic model.

\paragraph{Setup and study process.}
We hired $24$ participants, each of whom was given an outline of cultural learning and asked to talk to models based on the outline.
They can ask any related questions and express their opinions to models.
Afterwards, the participants took a cultural understanding exam of VSM 2013~\citep{vsm13,lamport94} which they had never come into contact.
We then computed the Euclidean distance between the ground truth and their answers from six cultural dimensions (rf. \Cref{sec-exp-align}).
For comparison, $12$ participants learned with our models and the other $12$ learned with GPT-4 to study $6$ cultures: Arabic, Bengali, German, Korean, Portuguese, and Spanish culture.
Each culture was learned by four participants, two of them learning with our models and the others learning with GPT-4.
Detailed information on the participants can be found in \cref{tb-human-participant}.
During the study process, first, each participant was given an outline for cultural learning written by cultural experts.
The outline (\Cref{sec-append-exp-education}), serves as the guideline for efficient learning.
Then, we asked the participants to freely communicate with the models to learn about specific cultures.
After the examination, we asked the participants to give a score of 1-5 to indicate their satisfaction with the learning process.
In this study, our aim was to answer two questions: 1) What is the learning performance of the participants with our models and GPT-4? 2) How are their learning experience?

\begin{wraptable}{r}{.45\textwidth}
\vspace{-.2in}
\caption{Results on situated learning.}
\vspace{-.1in}
\label{tb-human-result}
\resizebox{.45\textwidth}{!}{
\begin{tabular}{l|cc|cc}
\toprule
\multirow{2}{*}{Model} & \multicolumn{2}{c|}{Distance$\downarrow$} & \multicolumn{2}{c}{User experience$\uparrow$} \\ 
 & \multicolumn{1}{c}{GPT-4} & Ours & \multicolumn{1}{c}{GPT-4} & Ours \\ \midrule
Arabic & \multicolumn{1}{c}{89.89} & \textbf{69.57} & \multicolumn{1}{c}{4} & \textbf{5} \\ 
Bengali & \multicolumn{1}{c}{339.84} & \textbf{304.54} & \multicolumn{1}{c}{3} & \textbf{5} \\ 
Germany & \multicolumn{1}{c}{224.68} & \textbf{173.12} & \multicolumn{1}{c}{2} & \textbf{3} \\ 
Korean & \multicolumn{1}{c}{222.39} & \textbf{183.62} & \multicolumn{1}{c}{2} & \textbf{4} \\ 
Spanish & \multicolumn{1}{c}{143.33} & \textbf{102.53} & \multicolumn{1}{c}{4} & \textbf{5} \\ 
Turkish & \multicolumn{1}{c}{273.43} & \textbf{221.12} & \multicolumn{1}{c}{3} & \textbf{4} \\ \midrule
AVG & \multicolumn{1}{c}{215.59} & \textbf{175.75} & \multicolumn{1}{c}{3} & \textbf{4.33} \\ \bottomrule
\end{tabular}
}
\end{wraptable}

\paragraph{Results.}
\Cref{tb-human-result} shows the averaged results of different participants.
We have the following findings.
First, participants learning with our models achieved better performance in cultural examination than those with GPT-4 in all cultures.
This indicates that our fine-tuned models have a better cultural understanding than GPT-4.
Second, participants are more satisfied with communicating with our models than GPT-4.
Furthermore, many participants expressed that the responses from GPT-4 are vague. Even though we have prompted GPT-4 to be like a person from a specific culture, it always responds with neutral words that have no clear opinions or ideas. Instead, our models can provide straightforward opinions.

\section{Discussion}
\label{sec-discussion}

\subsection{Why \method benefits fine-tuning?}
\label{sec-discuss-why}

We analyze the effectiveness of \method in benefiting cultural model fine-tuning from two different aspects: Communication vs. direct generation of \llms and diversity of the generated data.

\begin{figure}[t!]
  \centering
  \subfigure[Benefits of communication.]{
    \includegraphics[width=.15\textwidth]{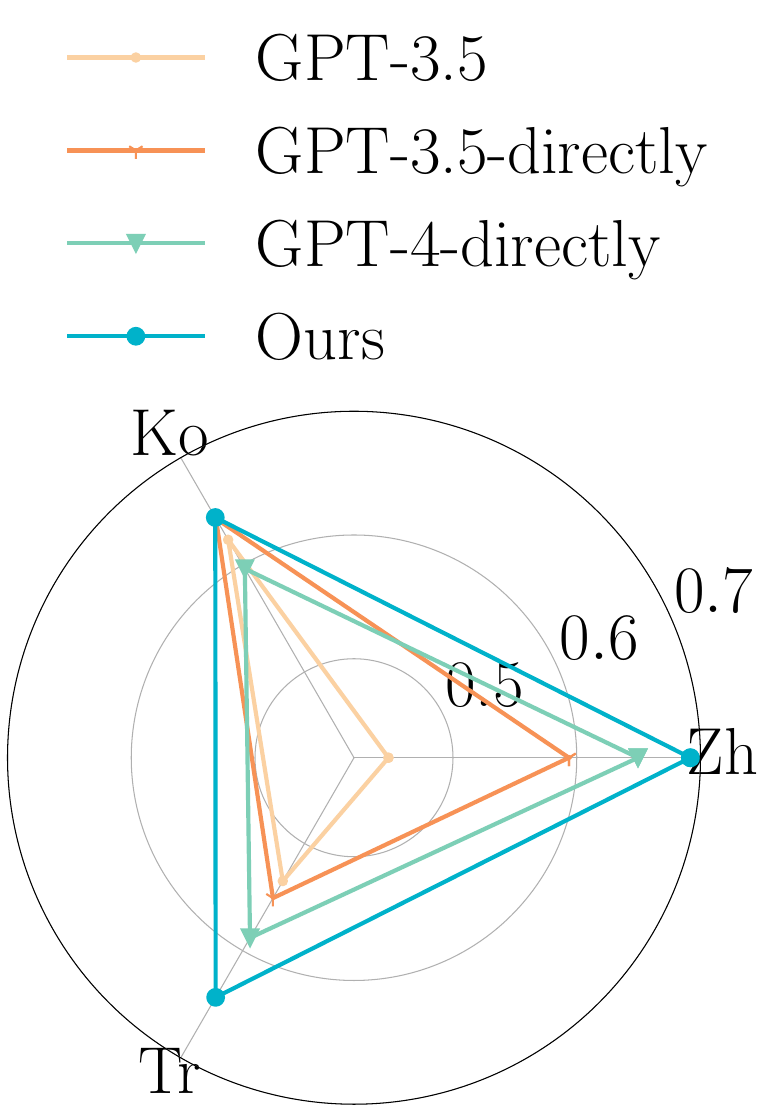}
    \label{fig-directly}
  }
  \hfill
  \subfigure[Diversity gain vs. CultureLLM~\citep{li2024culturellm}.]{
    \includegraphics[width=.25\textwidth]{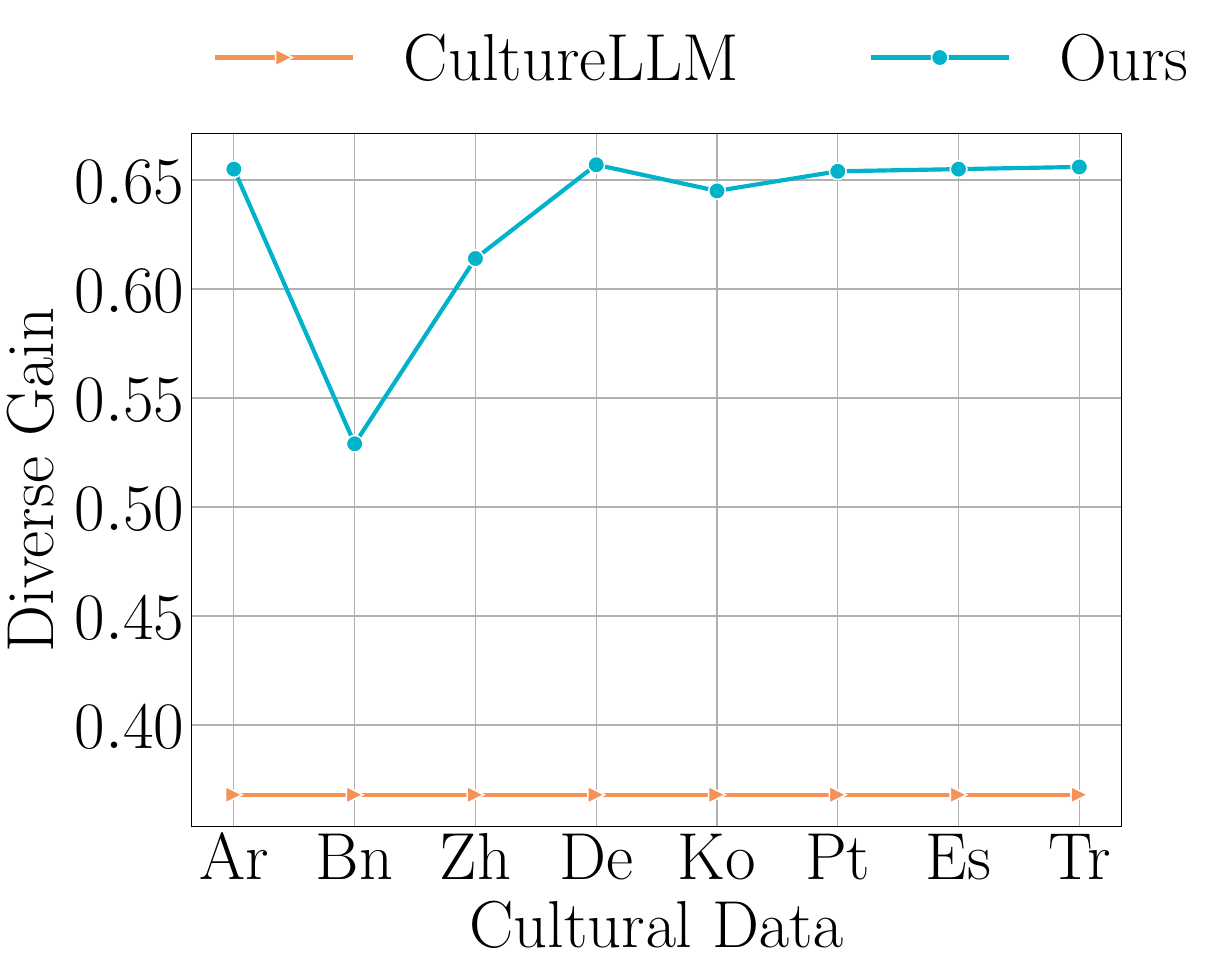}
    \label{fig-diverse-gain}
  }
  \hfill
  \subfigure[Influence of cultural background and gender.]{
    \includegraphics[width=.23\textwidth]{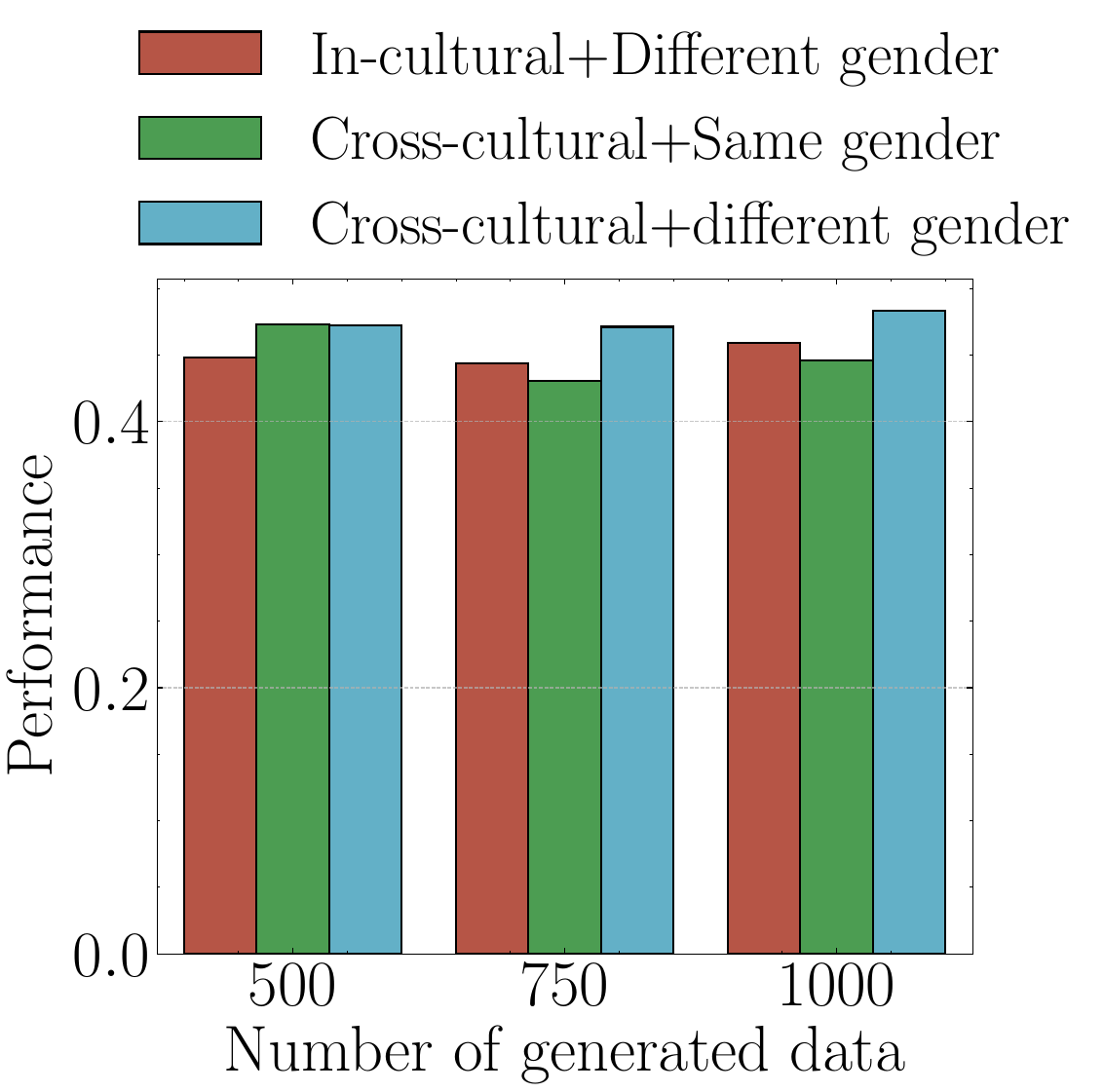}
    \label{fig-gender-culture}
  }
  \hfill
  \subfigure[Performance on BBH.]{
    \includegraphics[width=.22\textwidth]{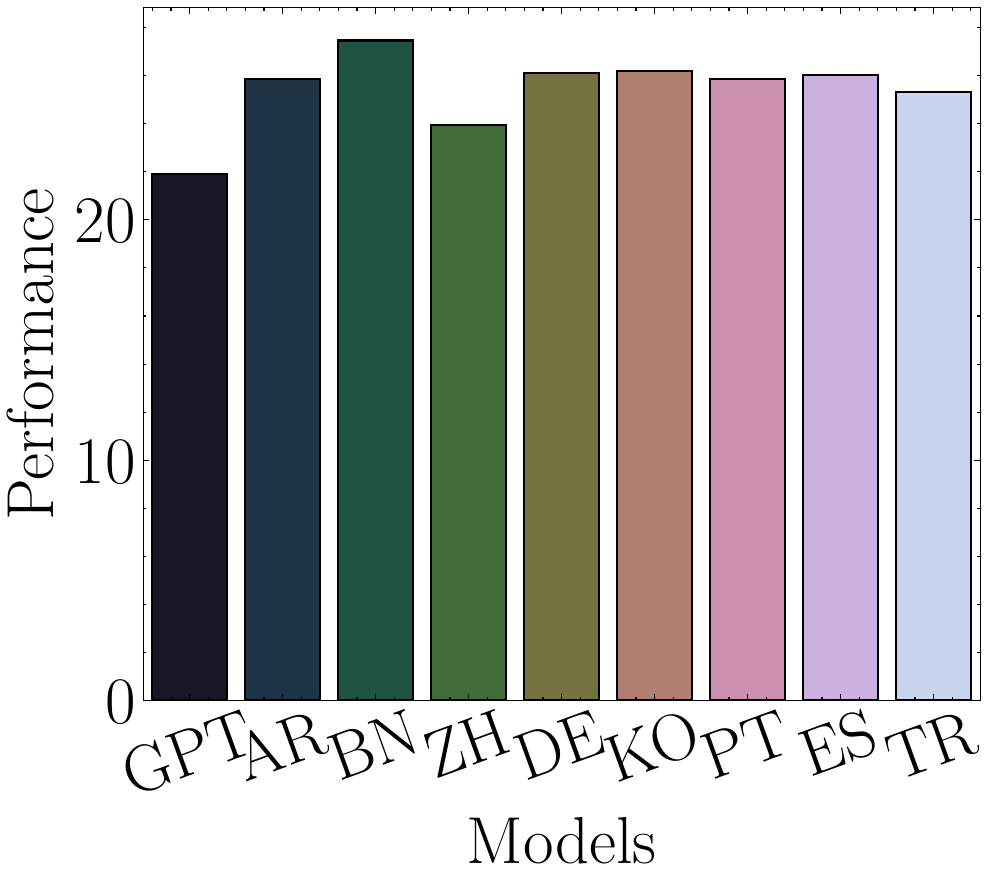}
    \label{fig-forget-BBH}
  }
  \vspace{-.1in}
  \caption{More discussions on \method.}
  \vspace{-.1in}
\end{figure}

\textbf{Cross-cultural communication vs. direct generation from GPT-4 / GPT-3.5.}
We compared the results of fine-tuning using data directly generated by GPT models (i.e., no communication).
We generated such data by prompting GPT-4 or GPT-3.5-turbo as: ``\prompt{Question: \{input\} Answer: \{output\} Please list 10 reasons to support the answer and number them}''.
Then, these data are used to fine-tune GPT-3.5-turbo.
\Cref{fig-directly} shows the performance on content moderation tasks in Chinese, Korean and Turkish cultures.
We see that data directly generated from GPT-4 is better than that from GPT-3.5, while our GPT-3.5-based models can outperform both of them.

\textbf{Diversity of the generated data.}
We also analyzed the diversity gain~\citep{bilmes2022submodularity} of the generated data for quality evaluation.
We compared with CultureLLM~\citep{li2024culturellm} and presented the results in \Cref{fig-diverse-gain}.
It indicates that \method can generate more diverse and high-quality data.

\subsection{Exploring agents' cultural background and gender}
\label{sec-dis-culture-gender}

To explore the influence of agent's cultural background and gender, we conducted three types of multi-agent communications in Arabic culture: ``In-cultural+Different gender'', ``Cross-cultural+Same gender'', and ``Cross-cultural+Different gender''.\footnote{For a fair comparison, the number of seed data for this part is $50$ from WVS~\citep{wvs} following CultureLLM~\citep{li2024culturellm}.}
For each setting, we fine-tuned three different models, whose training data is $500$, $750$, and $1000$, respectively.
We evaluated the performance of the models on content moderation tasks and presented the results in \Cref{fig-gender-culture}.
``Cross-cultural+Different gender'' exhibits the best performance and ability to generate more high-quality data.
This indicates the necessity of bringing more diversity in data generation, as conducted in \method.

\begin{wrapfigure}{r}{0.5\textwidth}
\vspace{-.2in}
  \begin{center}
    \includegraphics[width=.5\textwidth]{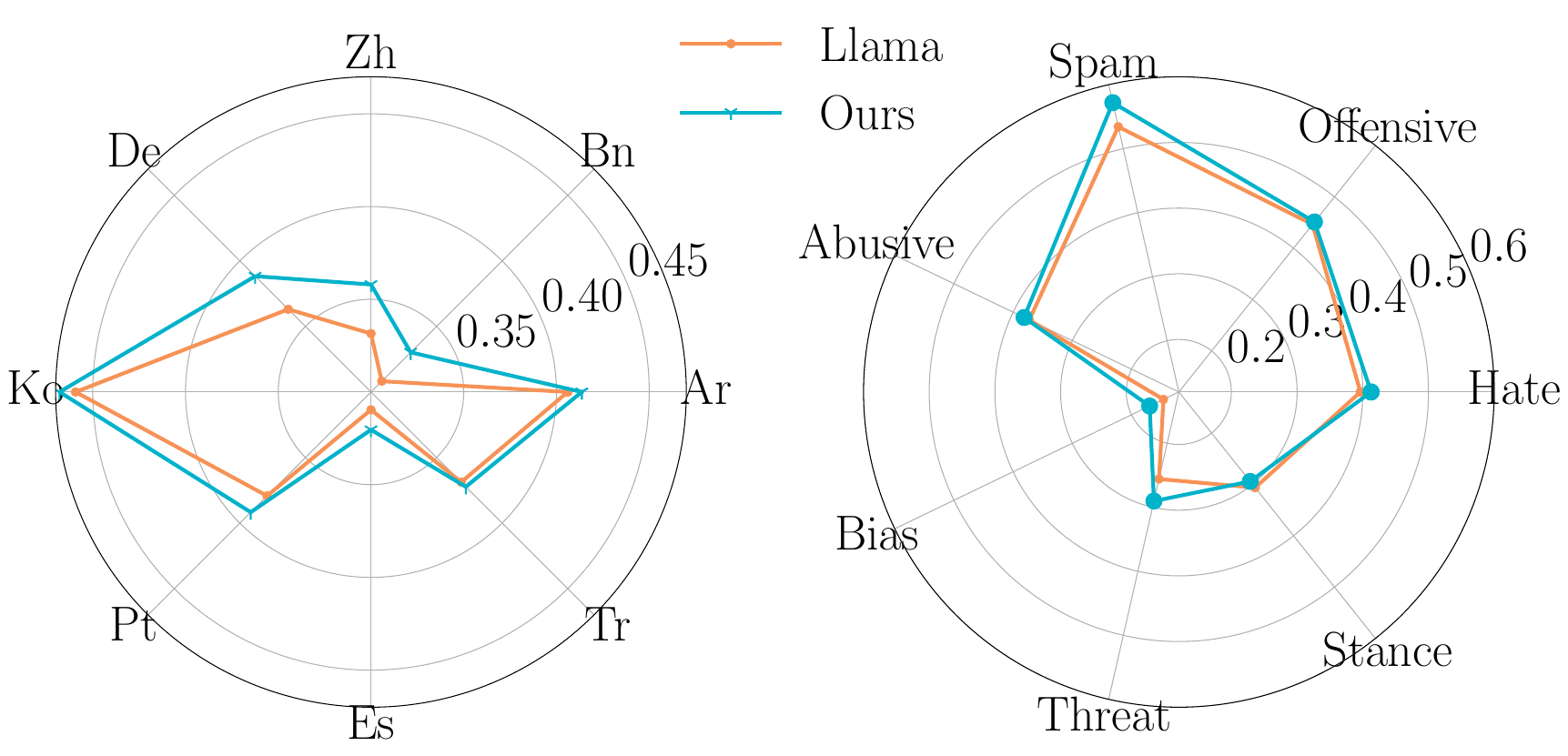}
  \end{center}
  \vspace{-.1in}
  \caption{Results of \method-Llama on content moderation for different cultures.}
  \vspace{-.2in}
  \label{fig-llama-result}
\end{wrapfigure}

\subsection{Fine-tuning vs. forgetting}

A potential dilemma arises when fine-tuning a large language model for specific tasks, as it may result in catastrophic forgetting of its original capabilities.
This section explores the extent of forgetting exhibited by \method on BIG-Bench-Hard (BBH)~\citep{suzgun2022challenging}, which comprises 21 tasks that assess semantic understanding and logical reasoning.
For cost efficiency, we sampled $100$ samples from each BBH task.
We evaluated our models against the baseline model, GPT-3.5-turbo.
The results in \Cref{fig-forget-BBH} indicate that \method generally maintains or even improves performance on most benchmarks, including the 21 tasks in BBH.
This improvement suggests potential latent relationships between cultural data and general benchmarks, implying that fine-tuning of cultural data could improve general reasoning abilities.

\subsection{Open-source fine-tuning with Llama2-70b}
\label{sec-discuss-llama}

To verify the generalization ability of our framework, we leveraged the generated data to fine-tune cultural-specific Llama-2-70b models and evaluate on content moderation tasks.
As shown in \Cref{fig-llama-result}, our models outperform Llama-2-70b in all $8$ cultures, especially in German, Chinese, Bengali and Portuguese cultures, which cover both low- and high-resource cultures. Furthermore, our models are also excellent in all $7$ tasks.\footnote{The detailed hyperparameter setting is shown in \Cref{tb-finetune}. We only conduct this pilot study to show that Llama2 also works in \method. The results might be further improved with more GPU resources.}
This verifies the generalization of \method as an effective data collection platform.

\section{Conclusions, Societal Impact, and Limitations}
\label{sec-conclusion}

This paper introduced \method, an LLM-powered multi-agent framework for cultural data collection through multi-agent communication.
\method can generate high-quality and diverse cross-cultural dialogue, which can be used to fine-tune culturally specific \llms.
We evaluated \method across three downstream tasks: content moderation, cultural alignment, and cultural education, indicating great improvement over GPT-4.

\method enhances fairness and inclusivity, reduces discrimination, and ensures accurate cultural representation. It improves global communication, fosters cross-cultural understanding, and supports multilingual societies. It benefits as bias-free LLMs build trust and align with responsible principles. Economically, it expands market reach and drives innovation. Social harmony improves by reducing stereotypes and preserving cultural heritage. It also aids compliance with anti-discrimination laws and supports inclusive education, promoting cultural awareness. Addressing cultural biases in LLMs creates more just, reliable, and beneficial AI systems, contributing to a more equitable world.

Our work has the following limitations.
1) More experiments can be performed by replacing GPT-3.5-Turbo in \method to discover more results.
2) Our fine-tuned models are mostly for high-resource cultures. The reason is that the dataset and benchmark on low-resource cultures are rare, and we can not find enough data for fine-tuning and evaluation.
3) More efficient fine-tuning techniques can be studied to support the fine-tuning of culturally specific \llms.

\section*{Disclaimer}

The human study was conducted following local laws and regulations, and the evaluation process was controlled to ensure that no irresponsible content was generated.
The authors respect all cultures studied in the world.
The results of the paper may change due to the change in OpenAI API and its model versions.

\bibliographystyle{plainnat}
\bibliography{refs}





\newpage
\appendix
\etocdepthtag.toc{mtappendix}
\etocsettagdepth{mtchapter}{none}
\etocsettagdepth{mtappendix}{subsection}
\tableofcontents

\section{Discussion on the Relationship between Culture and Language}

We strongly agree that language is \textbf{not} equal to, but only \textbf{a part of} culture. But using language to study culture is possible due to the following aspects:
\begin{enumerate}
    \item Existing literature on culture understanding shows that culture boundaries are fluid, dynamic and uncertain. Delanoy emphasizes that cultures are not homogeneous or static entities but are fluid and dynamic. He critiques essentialist views that rigidly define cultural boundaries and instead promotes a more nuanced understanding that considers the intersections of various cultural factors, such as ethnicity, language, religion, and socio-economic conditions~\citep{delanoy2020culture}. Appadurai also discusses the fluidity of cultural boundaries and the creation of new cultural forms~\citep{appadurai1996modernity}. Cultural boundaries can be geographical regions, language, religion and so on. Based on above statements, using language as cultural boundaries is reasonable.
    \item Existing NLP works on culture also leverage labguage as culture boundaries. \citet{naous2023having} focuses on Arabic and English culture. ~\citet{wang2023not} focuses on 8 different cultures: English, Chinese, French, Russian, German, Arabic, Japanese and Korean. \citet{liu2023multilingual} also use language to split different cultures. The authors work on English, German, Russian, Bengali, Chinese, and Indonesian culture. \citet{myung2024blend} is a hand-crafted benchmark for evaluate diverse cultures. They also use languages as culture boundaries.
    \item Most downstream benchmarks are classified via language and we cannot get more fine-grained perspectives. For example, if we want to evaluate the performance of Arabic model, we can find benchmarks in Arabic culture. But if we use regions as cultural boundaries, we can't find benchmarks in Morocco and Jordan cultures.
\end{enumerate}

\section{Details on the cross-cultural dialogue}

\subsection{Examples on seed data}

\begin{figure}[htbp]
    \centering
    \includegraphics[width=.80\textwidth]{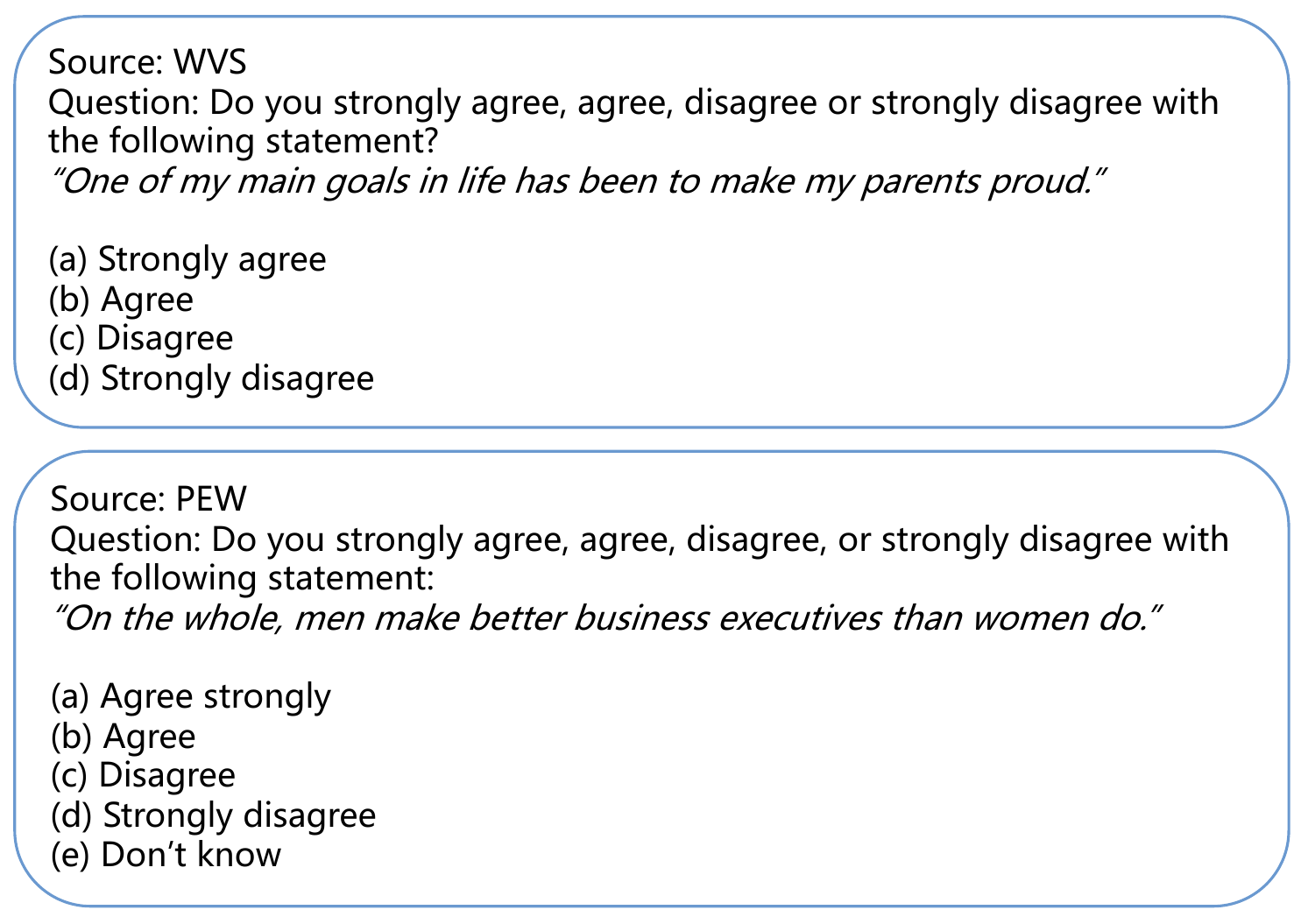}
  \vspace{-.1in}
  \caption{Example questions from the WVS and Pew explore perspectives on globally relevant political and ethical issues. Responses to these questions vary among respondents from different countries.}
  \label{fig-seed-data}
\end{figure}

\Cref{fig-seed-data} shows example questions from the WVS and Pew to explore perspectives on globally relevant political and ethical issues. Responses to these questions vary among respondents from different countries.

\subsection{The dialogue dataset}
\label{sec-append-dialogue}

\begin{figure}[htbp]
    \centering
    \includegraphics[width=.40\textwidth]{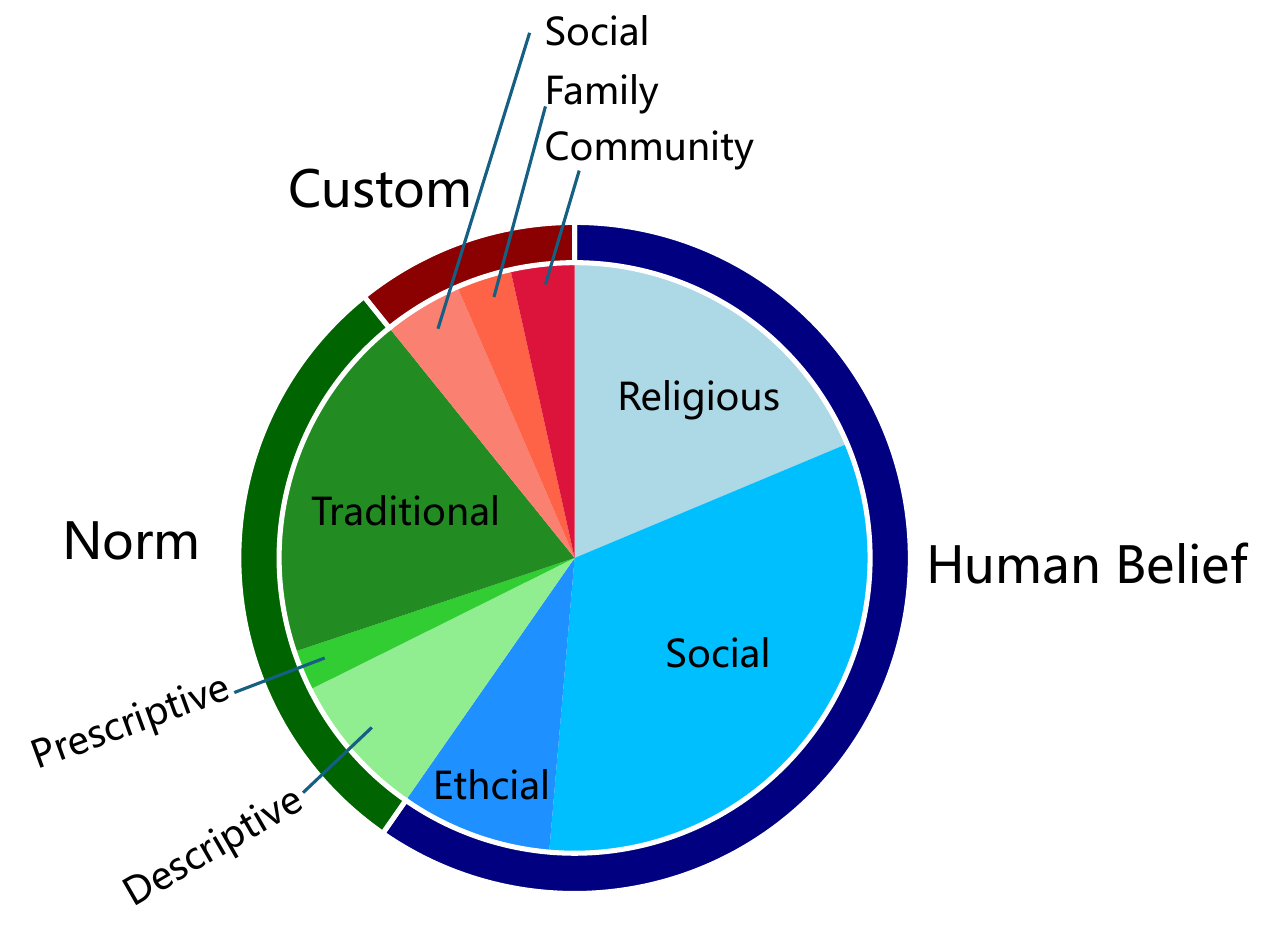}
  \vspace{-.1in}
  \caption{After multi-turn communications, we get a cross-cultural dialogues dataset (CCD), which involves data on $8$ different cultures. CCD contains human belief (59.68\%), norm (29.54\%), and custom (10.78\%).}
  \label{fig-dataset}
\end{figure}

\Cref{fig-dataset} shows the topic distribution of the generated cross-cultural dialogue dataset: human belief (59.68\%), norm (29.54\%), and custom (10.78\%). For human belief, there are three main types: religious beliefs (31.31\%), social beliefs (54.77\%), and ethical beliefs (13.92\%). For norm, the data can be divided into descriptive norms (26.98\%), prescriptive norms (7.48\%), and traditional norms (65.53\%). For custom, they can be classified into social customs (39.75\%), family customs (27.95\%), and community customs (32.30\%).

Then, we randomly sampled $750$ dialogues for each culture and evaluated the communication using prompts in \Cref{sec-append-prompt}.
As shown in \Cref{tb-observation}, on average, the average ratio of statements that express cross-cultural understanding is $80.80$\%.
The analysis also verifies the effectiveness of \method in extending topics and cross-cultural understanding.

\subsection{Details on data refinement}
\label{sec-append-refinement}

\begin{figure}
    \centering
    \includegraphics[width=\linewidth]{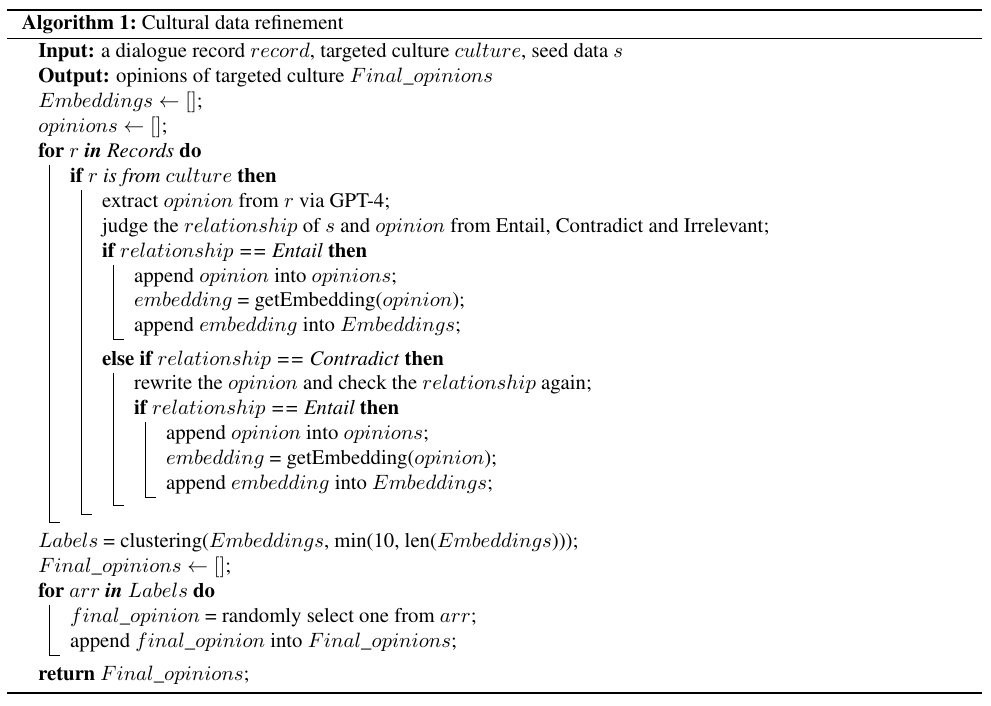}
    \caption{Pipeline of data refinement.}
    \label{algo-aug}
\end{figure}

Algorithm \Cref{algo-aug} shows the algorithm pipeline of data refinement.


\begin{table}[]
\centering

\caption{Interesting observation in Cross-cultural Dialogues. "Extend Rate" represents the ability of entending the topic. "Understanding / Others" represents the ratio of cross-cultural understanding statements and others.}
\label{tb-observation}
\resizebox{.8\textwidth}{!}{
\begin{tabular}{llllllllll}
\toprule
\multicolumn{1}{c}{}                                  & \multicolumn{1}{c}{Ar}    & \multicolumn{1}{c}{Bn}    & \multicolumn{1}{c}{Zh}    & \multicolumn{1}{c}{De}    & \multicolumn{1}{c}{Ko}    & \multicolumn{1}{c}{Pt}    & \multicolumn{1}{c}{Es}    & \multicolumn{1}{c}{Tr}    & \multicolumn{1}{c}{AVG}   \\ \midrule
\multicolumn{1}{c}{Extend rate}                       & \multicolumn{1}{c}{34.36} & \multicolumn{1}{c}{33.33} & \multicolumn{1}{c}{34.74} & \multicolumn{1}{c}{37.82} & \multicolumn{1}{c}{32.82} & \multicolumn{1}{c}{35.13} & \multicolumn{1}{c}{35.90} & \multicolumn{1}{c}{32.69} & \multicolumn{1}{c}{34.60} \\
\makecell{Understanding ratio} & \multicolumn{1}{c}{75.68} & \multicolumn{1}{c}{80.56} & \multicolumn{1}{c}{80.97} & \multicolumn{1}{c}{80.14} & \multicolumn{1}{c}{85.27} & \multicolumn{1}{c}{80.97} & \multicolumn{1}{c}{81.40} & \multicolumn{1}{c}{81.40} & \multicolumn{1}{c}{80.80} \\ \bottomrule
\end{tabular}
}
\end{table}

\section{Details on the test sets}
\label{sec-append-dataset}

The statistics of the datasets are shown in \cref{tb-datasets}, and we provide detailed instructions for them in the following.

\begin{table*}[htbp]

\caption{A brief introduction of the $8$ evaluation tasks and $41$ datasets. We list both the name and the size of test sets. For instance, ``OSACT5(2541)~\citep{mubarak2022overview}'' denotes that there are 2541 test samples in the dataset OSACT5.}
\label{tb-datasets}
\resizebox{\textwidth}{!}{
\begin{tabular}{c|p{2.8cm}|c|c}
\toprule
Culture & Country \& Territory & Task \& Dataset & \#Sample \\ \midrule
\makecell{Arabic\\(CulturePark-Ar)}  & Middle East & \makecell{\textit{Offensive language detection:} OSACT5(2541)~\citep{mubarak2022overview}. \\\textit{Hate detection:} Multi-Platform(675)~\citep{chowdhury2020multi}, \\OSACT5(2541)~\citep{mubarak2022overview}, and \\OSACT5\_finegrained(2541)~\citep{mubarak2022overview}. \\\textit{Vulgar detection:} Multi-Platform(675)~\citep{chowdhury2020multi}}    & 8,973 \\ \midrule
\makecell{Bangli\\(CulturePark-Bn)}  & Bangladesh & \makecell{\textit{Offensive language detection:} TRAC2020 Task1(1000)~\citep{trac2-dataset}, \\TRAC2020 Task2(1000)~\citep{trac2-dataset}. \\\textit{Hate detection:} Hate Speech(1000)~\citep{romim2021hate}. \\\textit{Threat detection:} BACD(1000)~\citep{Bangla-Abusive-Comment-Dataset}. \\\textit{Bias detection:} BACD(1000)~\citep{Bangla-Abusive-Comment-Dataset}.}     & 5,000 \\ \midrule
\makecell{Chinese\\(CulturePark-Zh)}  & China & \makecell{\textit{Spam detection:} CCS(1000)~\citep{jiang2019detect}. \\\textit{Bias detection:} CDial-Bias(1000)~\citep{zhou2022towards}.}  & 2,000 \\ \midrule
\makecell{German\\(CulturePark-De)}  & Germany and parts of Europe & \makecell{\textit{Offensive language detection:} GermEval2018(3531)~\citep{wiegand2018overview}. \\\textit{Hate detection:} IWG\_1(469)~\citep{ross2016hatespeech}, IWG\_2(469)~\citep{ross2016hatespeech}, \\HASOC2020(850)~\citep{HASOC2020}, \\and multilingual-hatecheck(1000)~\citep{rottger2022multilingual}.}  & 6,319 \\ \midrule
\makecell{Korean\\(CulturePark-Ko)}  & South Korea & \makecell{\textit{Hate detection:} hateSpeech(1000)~\citep{moon-etal-2020-beep}, \\and HateSpeech2(1000)~\citep{Korean-HateSpeech-dataset}. \\\textit{Abusive detection:} AbuseEval(1000)~\citep{caselli2020feel}, \\CADD(1000)~\citep{song2021large}, \\and Waseem(1000)~\citep{waseem2016hateful}.}  & 5,000 \\ \midrule
\makecell{Portuguese\\(CulturePark-Pt)} & Brazil and parts of Latin America & \makecell{\textit{Offensive language detection:} OffComBR(1250)~\citep{Pelle2017}, \\and HateBR(1000)~\citep{vargas-etal-2022-hatebr}. \\\textit{Bias detection:} ToLD-Br-homophobia(1000)~\citep{leite2020toxic}, \\and ToLD-Br-misogyny(1000)~\citep{leite2020toxic}. \\\textit{Abusive detection:} ToLD-Br-insult(1000)~\citep{leite2020toxic}.}  & 5,250 \\ \midrule
\makecell{Spanish\\(CulturePark-Es)} & Argentina, Mexico, and parts of Latin America & \makecell{\textit{Offensive language detection:} AMI(1000)~\citep{fersini2018overview}, \\MEX-A3T(1000)~\citep{alvarez2018overview}. \\\textit{Hate detection:} HatEval 2019(1000)~\citep{basile2019semeval}. \\\textit{Bias detection:} DETOXIS\_stereotype(1000)~\citep{de2021ai}, and \\DETOXIS\_improper(1000)~\citep{de2021ai}. \\\textit{Abusive detection:} DETOXIS\_abusive(1000)~\citep{de2021ai}, \\DETOXIS\_mockery(1000)~\citep{de2021ai}. \\\textit{Aggressive detection:} DETOXIS\_aggressiveness(1000)~\citep{de2021ai}. \\\textit{Stance detection:} DETOXIS\_stance(1000)~\citep{de2021ai}.}  & 9,000 \\ \midrule
\makecell{Turkish\\(CulturePark-Tr)} & Turkey & \makecell{\textit{Offensive language detection:} SemEval-2020(3528)~\citep{zampieri2020semeval}, \\and offenseKaggle\_2(1000)~\citep{turkish-offensive-language-detection}. \\\textit{Abusive detection:} ATC(1000)~\citep{karayiugit2021detecting}. \\\textit{Spam detection:} Turkish Spam(825)~\citep{misc_turkish_spam_v01_530}. \\\textit{Fine-grained offensive detection:} offenseCorpus(1000)~\citep{coltekin2020lrec}.}  & 7,353 \\ \bottomrule                                                 
\end{tabular}
}
\end{table*}

\subsection{Arabic}

OffenseEval2020~\citep{zampieri2020semeval} dataset was created to address the issue of offensive language in social media. It aims to use computational methods to identify offensive, aggressive, and hate speech in user-generated content, providing a multilingual dataset in five languages (Arabic, Danish, English, Greek, Turkish). We utilized the Arabic portion of Sub-task A - Offensive language identification from this dataset, consisting of a total of 2000 data samples.

OSCAT4~\citep{husain2005osact4} dataset aims to detect and categorize offensive language in Arabic tweets, with two sub-tasks: detecting if a post is offensive or not, and identifying the offensive content type as hate speech or not hate speech. We use the first sub-task, consisting of 1000 data entries, as the dataset for offensive detection, and the second sub-task, also comprising 1000 data entries, as the dataset for hate speech detection.

Multi-Platform~\citep{chowdhury2020multi} dataset is a collection of 4000 comments in Dialectal Arabic from social media platforms, focusing on offensive language. It is intended for studying offensive language in news comments published by international news organizations. We utilized a total of 1000 annotated data samples indicating whether they are offensive and 675 annotated data samples indicating whether they are vulgar.

OSACT5~\citep{mubarak2022overview} dataset consists of 12,698 Arabic tweets collected between June 2016 and November 2017, labeled for offensiveness and fine-grained hate speech types using emojis commonly found in offensive communications, providing a resource for offensive and hate speech detection and classification tasks. The dataset consists of three subtasks: offensiveness detection, hate speech detection, and fine-grained hate speech detection. We utilized 2,541 data samples for each of these tasks.

ASHT~\citep{kaddoura2024dataset} dataset contains 132,421 Arabic tweets collected from Twitter, classified as either ham (non-spam) or spam, providing a valuable resource for researchers in Arabic natural language processing (NLP) and serving as a benchmark for research in Arabic NLP, cybersecurity, data science, and social network analysis. We utilized a subset of 1,000 data samples for the spam detection section.

\subsection{Bengali}

TRAC2020~\citep{trac2-dataset} dataset is a multilingual annotated corpus of social media comments, encompassing misogynistic and aggressive comments in Indian English, Hindi, and Indian Bangla. It consists of over 20,000 comments and is annotated at two levels - aggression (overtly aggressive, covertly aggressive, and non-aggressive) and misogyny (gendered and non-gendered). Baseline experiments were conducted to develop misogyny classifiers for the three languages. TRAC2020 consists of two tasks: Aggression Detection and Misogynistic Aggression Detection. We utilized 1,000 data samples for each of Task 1 and Task 2.

BAD~\citep{sharif2022tackling} dataset is a novel Bengali aggressive text dataset (called 'BAD') with two-level annotation, designed to identify and classify aggressive content in Bengali language. It achieves high accuracy through a weighted ensemble technique and outperforms other machine learning and deep learning baselines, with a weighted f1-score of 93.43\% for identification and 93.11\% for categorization tasks. We utilized a subset of one thousand data samples as the Offensive dataset.

Hate Speech~\citep{romim2021hate} dataset consists of 30,000 social media user comments, covering seven categories including sports, entertainment, religion, politics, crime, celebrities, TikTok, and memes. It has been annotated through crowdsourcing and expert validation for research purposes in detecting hate speech in Bengali language. The dataset also provides benchmark experimental results for multiple deep learning models and pre-trained Bengali word vectors. We utilized 1,000 data samples from the dataset for Hate Detection.

BACD~\citep{Bangla-Abusive-Comment-Dataset} dataset is a dataset for the Bengali language, consisting of a total of 10,200 data points with annotations for toxic, threat, obscene, insult, and racism labels. We utilized 1,000 data points from this dataset for Threat Detection and Bias Detection tasks respectively.

\subsection{Chinese}

CCS~\citep{jiang2019detect} dataset consists of two real-world spam datasets: one is an SMS dataset, and the other is a product review dataset. Both datasets were manually labeled by professionals as spam or regular emails, and their sizes and label distributions were summarized. We utilized 1000 data samples from this dataset for Spam Detection.

CDial-Bias~\citep{zhou2022towards} Dataset is the first annotated Chinese social bias dialog dataset, utilized to establish a benchmark for measuring dialog bias and evaluate Chinese generative models for social bias presence. We utilized 1000 data samples from it for bias detection.

CValues~\citep{xu2023cvalues} is a Chinese human values evaluation benchmark that measures the alignment ability of large language models in terms of safety and responsibility, providing both manual and automatic evaluation to assess their performance and identify areas for improvement. We utilized 1712 data samples from the dataset for Stance detection.

\subsection{Germany}

GermEval2018~\citep{wiegand2018overview} dataset is used for identifying offensive language in German tweets, including both coarse-grained binary classification tasks and fine-grained multi-class classification tasks. We used 3,531 data points for Offensive Detection.

IWG~\citep{ross2016hatespeech} dataset aims to assess the feasibility of reliably annotating hate speech and explore the consistency between existing definitions and subjective ratings. The results indicate low reliability in users' judgments of hate speech, suggesting a need for more detailed annotation instructions. Each data instance in the dataset was annotated by two experts, and we selected 469 instances with annotations from both experts for Hate Detection, denoted as IWG\_1 and IWG\_2 respectively.

HASOC2020~\citep{HASOC2020} dataset is a multilingual research forum and data challenge that offers tasks for identifying problematic content in English, German, and Hindi. It consists of over 10,000 annotated tweets from Twitter, and includes both coarse-grained and fine-grained classification tasks. We utilized a subset of 850 German language data from the HASOC dataset for Hate Detection.

Multilingual HateCheck~\citep{rottger2022multilingual} is a comprehensive dataset of functional tests for hate speech detection models in ten languages, addressing the need for more effective models and uncovering critical weaknesses for monolingual and cross-lingual applications. We utilized 1000 data points from the German section of the dataset for Hate Detection.

\subsection{Korean}

K-MHaS~\citep{lee-etal-2022-k} is a multi-label dataset consisting of 109k utterances from Korean news comments, designed for hate speech detection. It effectively handles Korean language patterns, provides multi-label classification with 1 to 4 labels, and considers subjectivity and intersectionality. Strong baseline experiments using Korean-BERT-based language models show that KR-BERT with a sub-character tokenizer performs the best by recognizing decomposed characters in each hate speech class. We utilized 1000 data samples from the dataset for Hate Detection.

HateSpeech~\citep{moon-etal-2020-beep} dataset is a collection of 9.4K manually labeled entertainment news comments in Korean, aimed at identifying toxic speech, social bias, and hate speech. It provides benchmarks using CharCNN, BiLSTM, and BERT models, with BERT achieving the highest performance. The dataset is made publicly available and open for competition. We utilized 1000 data samples from the dataset for Hate Detection.

HateSpeech2~\citep{Korean-HateSpeech-dataset} dataset was created by the Natural Language Processing Laboratory (NLP) at Korea National University and it includes the original dataset, a vocabulary of offensive language, annotations, and dataset examples. The dataset is used for labeling malicious comments and has been built with word embeddings. We utilized 1000 data samples from the dataset for Hate Detection.

AbuseEval~\citep{caselli2020feel} is a newly created dataset that addresses issues in annotating offensive and abusive language, specifically considering the degree of explicitness, target presence, and contextual interaction across different abusive language phenomena. We utilized 1000 data samples from the dataset for Abusive Detection.

CADD~\citep{song2021large} is a comprehensive dataset for detecting abusive language in English Reddit posts, featuring multifaceted labels and contextual information, collected through large-scale crowdsourcing and yielding meaningful performance with state-of-the-art language models. We utilized 1000 data samples from the dataset for Abusive Detection.

Waseem~\citep{waseem2016hateful} dataset, based on critical race theory, provides annotations for over 16k tweets and aims to detect hate speech on social media by analyzing linguistic features, extra-linguistic features, and a dictionary of the most indicative words in the data. We utilized 1000 data samples from the dataset for Abusive Detection.

\subsection{Portuguese}

OffComBR~\citep{Pelle2017} dataset is an annotated collection of offensive comments in Portuguese, gathered from news comment sections on the Brazilian web. It serves the purpose of classifying user-generated text as either positive or negative, providing a baseline for future research on the topic of hate speech detection in Portuguese. We utilized 1250 data samples from this dataset for offensive detection.

HateBR~\citep{vargas-etal-2022-hatebr} dataset is the first large-scale expert annotated corpus of Brazilian Instagram comments, specifically collected from politicians' accounts, providing binary/offensiveness-level classification and nine hate speech groups, outperforming the current state-of-the-art for Portuguese language offensive language and hate speech detection. We utilized 1000 data samples from this dataset for offensive detection.

ToLD-Br~\citep{leite2020toxic} is a large-scale dataset for Brazilian Portuguese, consisting of annotated tweets categorized as toxic or non-toxic, aiming to detect and prevent the proliferation of toxicity in social media, addressing the need for multilingual approaches and models aware of different categories of toxicity. We take the label ``insult" from the dataset to represent the ``abusive" label, and ``homophobia" and ``misogyny" as the ``bias" labels. We have selected 1000 data samples for Abusive Detection, 1000 samples for Bias Detection, and 1000 samples for Bias Detection.

\subsection{Spanish}

AMI~\citep{fersini2018overview} dataset is a collection of Spanish and English tweets used for identifying misogyny, categorizing misogynistic behavior, and classifying targeted individuals, with contributions from multiple teams and countries. We used 1000 Spanish language data for offensive detection.

MEX-A3T~\citep{alvarez2018overview} dataset, from the track at IberEval 2018, comprises Mexican Spanish tweets and focuses on two tasks: author profiling, which aims to identify the residence and occupation of Twitter users, and aggressiveness detection, to distinguish between aggressive and non-aggressive tweets. This dataset was created specifically for these tasks and was analyzed and compared in a paper discussing the participants' results. We used 1000 data samples for offensive detection.

OffendES~\citep{plaza2021offendes} dataset is a collection of 47,128 manually labeled Spanish comments from social media platforms, focusing on offensive language targeted at young influencers. It provides pre-defined offensive categories and includes confidence scores, enabling both multi-class classification and multi-output regression studies. We used 1000 data samples for offensive detection.

HatEval 2019~\citep{basile2019semeval} dataset focuses on detecting hate speech against immigrants and women in Spanish and English Twitter messages. It includes two classification tasks: identifying the presence of hate speech and distinguishing between individual and group targets. HatEval was a popular SemEval-2019 task with numerous submissions and participant system analysis. We used 1000 data samples for hate detection.

HaterNet~\citep{pereira2019detecting} dataset is an intelligent system used for monitoring and visualizing hate speech on Twitter. It provides a novel public dataset of Spanish hate speech, consisting of 6,000 expert-annotated tweets. We used 1000 data samples for hate detection.

DETOXIS~\citep{de2021ai} dataset is designed for the task of detecting toxic comments in online news discussions related to immigration. It includes toxicity detection and toxicity level detection. Participating teams achieved good results using the BERT model on this dataset. We classified them into tags such as stereotype, improper, abusive, mockery, aggressiveness, and stance, and selected 1000 data samples for each category for Bias detection, Abusive detection, Aggressiveness detection, and Stance detection.

\subsection{Turkish}

SemEval-2020~\citep{zampieri2020semeval} provided a new, large-scale semi-supervised training dataset of over nine million English tweets and expanded the task to include four new languages, allowing for cross-lingual training and analysis. We used 3528 data samples in Turkish for Offensive Detection.

OffenseCorpus~\citep{coltekin2020lrec} is a corpus of Turkish offensive language, comprising randomly sampled micro-blog posts from Twitter. It contains 36,232 tweets collected over an 18-month period from April 2018 to September 2019. We used 1000 data samples for Offensive Detection.

OffenseKaggle~\citep{turkish-tweets} Dataset is a collection of Turkish tweets from Twitter, with around 40\% of them containing offensive or vulgar content. We used 1000 data samples for Offensive Detection.

OffenseKaggle\_2~\citep{turkish-offensive-language-detection} dataset is an enhanced version of an existing offensive language research dataset, which has been expanded and annotated using contextual data mining techniques. It addresses the issue of class imbalance in existing studies and provides a more comprehensive and robust dataset for Turkish offensive language detection tasks. We used 1000 data samples for Offensive Detection.

ATC~\citep{karayiugit2021detecting} dataset is a publicly available dataset for detecting abusive Turkish comments on Instagram. It consists of 10,528 abusive and 19,826 non-abusive comments, with sentiment annotations at the sentence level. We used 1000 data samples for Offensive Detection.

Turkish Spam~\citep{misc_turkish_spam_v01_530} dataset contains both spam and normal emails written in Turkish. A total of 330 spam emails and 496 normal emails were collected from several personal accounts. We used 825 pieces of data for spam detection.

OffenseCorpus~\citep{coltekin2020lrec} dataset is a large collection of Turkish offensive language from Twitter micro-blog posts, annotated based on recent practices. It includes 36,232 randomly sampled tweets from April 2018 to September 2019, with 19\% containing offensive language. We used 1000 of the data for fine-grained offensive detection.

\subsubsection{Details on Fine-tuning}

\begin{table}[]
\caption{Details on Fine-tuning GPT-3.5-turbo using OpenAI API.}
\label{tb-finetune}
\centering
\begin{tabular}{lllllllll}
\toprule
\multicolumn{1}{c}{Model}  & \multicolumn{1}{c}{Ar} & \multicolumn{1}{c}{Bn} & \multicolumn{1}{c}{Zh} & \multicolumn{1}{c}{De} & \multicolumn{1}{c}{Ko} & \multicolumn{1}{c}{Pt} & \multicolumn{1}{c}{Es} & \multicolumn{1}{c}{Tr} \\ \midrule
\multicolumn{1}{c}{Epochs} & \multicolumn{1}{c}{12} & \multicolumn{1}{c}{6}  & \multicolumn{1}{c}{7}  & \multicolumn{1}{c}{4}  & \multicolumn{1}{c}{2}  & \multicolumn{1}{c}{3}  & \multicolumn{1}{c}{5}  & \multicolumn{1}{c}{2}  \\
\bottomrule
\end{tabular}
\end{table}

We adjust the number of epochs to find the better performance. \Cref{tb-finetune} shows the details.

\begin{table}[]
\caption{Information of agents in \method}
\label{tb-agent-name}
\resizebox{\textwidth}{!}{
\begin{tabular}{l|llllllll}
\toprule
       & Arabian & Bengali & Chinese & German     & Korean & Portuguese & Spanish & Turkish \\ \midrule
Male   & Abdul   & Aarav   & Wei     & Maximilian & Joon   & João       & Javier  & Mehmet  \\
Female & Fatima  & Ananya  & Lili    & Sophia     & Haeun  & Maria      & María   & Ayşe    \\ 
\bottomrule
\end{tabular}
}
\end{table}
\begin{table}[]
\caption{Number of generated data for different cultures}
\label{tb-data-num}
\resizebox{\textwidth}{!}{
\begin{tabular}{llllllllll}
\toprule
\multicolumn{1}{c}{}                             & \multicolumn{1}{c}{Arabic} & \multicolumn{1}{c}{Bengali} & \multicolumn{1}{c}{Chinese} & \multicolumn{1}{c}{German} & \multicolumn{1}{c}{Korean} & \multicolumn{1}{c}{Portuguese} & \multicolumn{1}{c}{Spanish} & \multicolumn{1}{c}{Turkish} & \multicolumn{1}{c}{Total} \\ \midrule
\multicolumn{1}{c}{\#Seed data}                  & \multicolumn{1}{c}{450}    & \multicolumn{1}{c}{650}     & \multicolumn{1}{c}{250}     & \multicolumn{1}{c}{550}    & \multicolumn{1}{c}{550}    & \multicolumn{1}{c}{550}        & \multicolumn{1}{c}{550}     & \multicolumn{1}{c}{550}     & \multicolumn{1}{c}{4100}  \\
\multicolumn{1}{c}{\#Generated data}             & \multicolumn{1}{c}{4500}   & \multicolumn{1}{c}{6500}    & \multicolumn{1}{c}{2500}    & \multicolumn{1}{c}{5500}   & \multicolumn{1}{c}{5500}   & \multicolumn{1}{c}{5500}       & \multicolumn{1}{c}{5500}    & \multicolumn{1}{c}{5500}    & \multicolumn{1}{c}{41000} \\
\bottomrule
\end{tabular}
}
\end{table}

\section{Details in experiments}
\label{sec-append-exp}

\subsection{Cultural alignment via Hofstede's cultural dimentions theory}
\label{sec-append-alignment}

The survey identified six dimensions of national culture: Power Distance Index (PDI), Individualism vs. Collectivism (IDV), Masculinity vs. Femininity (MAS), Uncertainty Avoidance Index (UAI), Long-Term Orientation vs. Short-Term Orientation (LTO) and Indulgence vs. Restraint (IND). VSM 2013 is an authoritative and famous cultural questionnaire devised by Hofstede that is used in \cite{masoud2023cultural,cao2023assessing}. In this experiment, we evaluate the cultural alignment of our models on $8$ cultures and compare with the state-of-the-art models: gpt-3.5-turbo and gpt-4.

To be specific, the VSM 2013 have $24$ questions in total. The computation of six cultural dimensions is based on the following formulas:
\begin{equation}
PDI=35(\mu_{Q7}-\mu_{Q2})+25(\mu_{Q20}-\mu_{Q23}) + C_{PDI}
\end{equation}
\begin{equation}
IDV=35(\mu_{Q4}-\mu_{Q1})+35(\mu_{Q9}-\mu_{Q6}) + C_{IDV}
\end{equation}
\begin{equation}
MAS=35(\mu_{Q5}-\mu_{Q3})+25(\mu_{Q8}-\mu_{Q10}) + C_{MAS}
\end{equation}
\begin{equation}
UAI=40(\mu_{Q18}-\mu_{Q15})+25(\mu_{Q21}-\mu_{Q24}) + C_{UAI}
\end{equation}
\begin{equation}
LTO=40(\mu_{Q13}-\mu_{Q14})+25(\mu_{Q19}-\mu_{Q22}) + C_{LTO}
\end{equation}
\begin{equation}
IVR=35(\mu_{Q12}-\mu_{Q11})+40(\mu_{Q17}-\mu_{Q16}) + C_{IVR}
\end{equation}

$\mu$ means the average of all the answers to each question. $C$ is constants that can be used to adjust to scores to fit a range between $0$ and $100$ or anchor new data to Hofstede's old dataset~\citep{lamport94}.

We get the Euclidean distance of the gaps from six cultural dimensions as the metric $Distance$. Larger $Distance$ means weak cultural understanding ability of \llms, and vice versa. 
\begin{equation}
\label{eq-distance}
Distance = \sqrt{\sum (d_{model} - d_{hofstede})^2}, \forall d \in \{PDI,IDV,MAS,UAI,LTO,IVR\}.
\end{equation}

\subsection{Details on situated learning}
\label{sec-append-exp-education}

The information of participants is shown in \Cref{tb-human-participant}.

\begin{table}[]
\label{tb-human-participant}
\caption{Information on participants in human study}
\centering
\begin{tabular}{c|cc|cc}
\toprule
\multicolumn{1}{c|}{Gender}    & \multicolumn{1}{c}{Male}     & \multicolumn{1}{c|}{12} & \multicolumn{1}{c}{Female} & \multicolumn{1}{c}{12} \\ \midrule
\multicolumn{1}{c|}{Education} & \multicolumn{1}{c}{Bachelor} & \multicolumn{1}{c|}{15} & \multicolumn{1}{c}{Master} & \multicolumn{1}{c}{9} \\ \midrule
\multicolumn{1}{c|}{Age}       & \multicolumn{1}{c}{22}       & \multicolumn{1}{c}{4} &                            &                        \\
                              & \multicolumn{1}{c}{23}       & \multicolumn{1}{c}{6} &                            &                        \\
                              & \multicolumn{1}{c}{24}       & \multicolumn{1}{c}{6} &                            &                        \\
                              & \multicolumn{1}{c}{25}       & \multicolumn{1}{c}{8}  &                            &                     \\ \bottomrule
\end{tabular}
\end{table}

\subsubsection{Details on training procedure}

The detailed outline is shown below:
\begin{enumerate}
    \item What factors do you prioritize when selecting a job, and what reasons underlie your choices?
    \item How do you weigh the significance of these aspects in your personal life: leisure time, consideration for others' wishes, assisting friends, and frugality? 
    \item What about your emotional well-being? Do feelings of nervousness or happiness play a significant role?
    \item What are your thoughts regarding your country and the individuals in your community?
    \item What about your state of health?
    \item Are you hesitant to disagree with your boss?
    \item Should a good manager have a precise answer to every question that a subordinate may raise about his or her work?
    \item How do you think about the relation between persistent efforts and results?
    \item Is it detrimental to have two bosses?
    \item Would you consider breaking the rules of a company or organization when they fail to align with your interests?
\end{enumerate}

The satisfaction scores are described as follows:
\begin{enumerate}[leftmargin=2em]
\setlength\itemsep{0em}
    \item Very Dissatisfied. Individuals feel significant disappointment and displeasure with the product or service. Expectations are not met to a substantial degree, often leading to feelings of frustration. This level typically results in complaints, negative reviews, or even discontinuation of use. Users are likely to advise others against the product or service.
    \item Dissatisfied. Users are somewhat unhappy and find that the product or service falls short of their expectations but not to a severe extent. They might tolerate the shortcomings temporarily but remain open to better alternatives. This level of dissatisfaction could prompt them to express concerns or seek improvements.
    \item Neutral. Users at this level neither feel particularly satisfied nor dissatisfied. Their expectations are met adequately but not impressively. There is no strong inclination either to complain or to commend. These users might continue using the service or product out of convenience rather than loyalty.
    \item Satisfied. Individuals are pleased with the product or service as it meets their expectations well. They experience a sense of fulfillment and value from their choice. While not overly enthusiastic, they are likely to continue using the product or service and may recommend it to others based on their positive experience.
    \item Very Satisfied. This level represents users who are exceptionally pleased with their experience. The product or service not only meets but exceeds their expectations, providing added value or delight. These users are highly likely to become repeat buyers, advocates, or even promoters of the brand, sharing their positive experiences through word of mouth or social media endorsements.
\end{enumerate}

\begin{table}[]
\caption{Prompts for evaluation.}
\label{tb-prompt-eval}
\centering
\resizebox{\textwidth}{!}{
\begin{tabular}{l m{16cm}}
\toprule
Task                           & Prompt                                                                                                                                                                                                                                                                                                                                                                                                                                                                                                        \\ \midrule
offensive\_detect              & if the following sentence is offensive, just say \textbackslash{}"OFF\textbackslash{}", otherwise, say \textbackslash{}"NOT\_OFF\textbackslash{}" without explanation:                                                                                                                                                                                                                                                                                                                                        \\
hate\_detect                   & if the following sentence has hate speech, just say \textbackslash{}"HS\textbackslash{}", otherwise, say \textbackslash{}"NOT\_HS\textbackslash{}" without explanation:                                                                                                                                                                                                                                                                                                                                       \\
vulgar\_detect                 & if the following sentence has vulgar speech, just say \textbackslash{}"Vulgar\textbackslash{}", otherwise, say \textbackslash{}"NOT\_Vulgar\textbackslash{}" without explanation:                                                                                                                                                                                                                                                                                                                             \\
spam\_detect                   & if the following sentence is spam tweet, just say \textbackslash{}"Spam\textbackslash{}", otherwise, say \textbackslash{}"NOT\_Spam\textbackslash{}" without explanation:                                                                                                                                                                                                                                                                                                                                     \\
stereotype\_detect             & if the following sentence has stereotype speech, just say \textbackslash{}"1\textbackslash{}", otherwise, say \textbackslash{}"0\textbackslash{}" without explanation:                                                                                                                                                                                                                                                                                                                                        \\
mockery\_detect                & if the following sentence has mockery speech, just say \textbackslash{}"1\textbackslash{}", otherwise, say \textbackslash{}"0\textbackslash{}" without explanation:                                                                                                                                                                                                                                                                                                                                           \\
insult\_detect                 & if the following sentence has insult speech, just say \textbackslash{}"1\textbackslash{}", otherwise, say \textbackslash{}"0\textbackslash{}" without explanation:                                                                                                                                                                                                                                                                                                                                            \\
improper\_detect               & if the following sentence has improper speech, just say \textbackslash{}"1\textbackslash{}", otherwise, say \textbackslash{}"0\textbackslash{}" without explanation:                                                                                                                                                                                                                                                                                                                                          \\
aggressiveness\_detect         & if the following sentence has aggressiveness speech, just say \textbackslash{}"1\textbackslash{}", otherwise, say \textbackslash{}"0\textbackslash{}" without explanation:                                                                                                                                                                                                                                                                                                                                    \\
toxicity\_detect               & if the following sentence has toxicity speech, just say \textbackslash{}"1\textbackslash{}", otherwise, say \textbackslash{}"0\textbackslash{}" without explanation:                                                                                                                                                                                                                                                                                                                                          \\
negative\_stance\_detect       & if the following sentence has negative stance speech, just say \textbackslash{}"1\textbackslash{}", otherwise, say \textbackslash{}"0\textbackslash{}" without explanation:                                                                                                                                                                                                                                                                                                                                   \\
homophobia\_detect             & if the following sentence has homophobia speech, just say \textbackslash{}"1\textbackslash{}", otherwise, say \textbackslash{}"0\textbackslash{}" without explanation:                                                                                                                                                                                                                                                                                                                                        \\
racism\_detect                 & if the following sentence has racism speech, just say \textbackslash{}"1\textbackslash{}", otherwise, say \textbackslash{}"0\textbackslash{}" without explanation:                                                                                                                                                                                                                                                                                                                                            \\
misogyny\_detect               & if the following sentence has misogyny speech, just say \textbackslash{}"1\textbackslash{}", otherwise, say \textbackslash{}"0\textbackslash{}" without explanation:                                                                                                                                                                                                                                                                                                                                          \\
threat\_detect                 & if the following sentence has threat speech, just say \textbackslash{}"1\textbackslash{}", otherwise, say \textbackslash{}"0\textbackslash{}" without explanation:                                                                                                                                                                                                                                                                                                                                            \\
bias\_on\_gender\_detect       & if the following speech expressing bias on gender, just say \textbackslash{}"1\textbackslash{}", otherwise, say \textbackslash{}"0\textbackslash{}" without explanation:                                                                                                                                                                                                                                                                                                                                      \\
hostility\_directness\_detect  & if the following speech expressing hostility directness, just say \textbackslash{}"1\textbackslash{}", otherwise, say \textbackslash{}"0\textbackslash{}" without explanation:                                                                                                                                                                                                                                                                                                                                \\
hate\_offens\_detect           & if the following sentence contains hate speech, just say \textbackslash{}"0\textbackslash{}", else if contains offensive language, say \textbackslash{}"1\textbackslash{}", otherwise, say \textbackslash{}"2\textbackslash{}" without explanation:                                                                                                                                                                                                                                                           \\
hate\_detect\_fine-grained     & if the following sentence doesn't have hate speech, just say \textbackslash{}"NOT\_HS\textbackslash{}", otherwise, label the hate speech with \textbackslash{}"HS1\textbackslash{}"(Race), \textbackslash{}"HS2\textbackslash{}"(Religion), \textbackslash{}"HS3\textbackslash{}"(Ideology), \textbackslash{}"HS4\textbackslash{}"(Disability), \textbackslash{}"HS5\textbackslash{}"(Social Class), \textbackslash{}"HS6\textbackslash{}"(Gender) without explanation:                                       \\
offensive\_detect\_finegrained & if the following sentence doesn't have offensive speech, just say \textbackslash{}"non\textbackslash{}", otherwise, label the offensive speech with \textbackslash{}"prof\textbackslash{}"(profanity, or non-targeted offense), \textbackslash{}"grp\textbackslash{}"(offense towards a group), \textbackslash{}"indv\textbackslash{}"(offense towards an individual), \textbackslash{}"oth\textbackslash{}"(ffense towards an other (non-human) entity, often an event or organization) without explanation: \\             \bottomrule                                    
\end{tabular}
}
\end{table}
\section{Prompts setting}
\label{sec-append-prompt}
\textbf{Prompts for statistical analysis}
\begin{enumerate}
    \item Do the two paragraphs discuss same topic? Just answer with Yes, or No.
    
    Paragraph 1: \{Paragraph 1\}
    
    Paragraph 2: \{Paragraph 2\}
    \item Does the paragraph reflect cross-cultural understanding? Just answer with Yes, or No.
    
    Paragraph: \{Paragraph\}
\end{enumerate}

\textbf{Prompts for evaluation}
For evaluation, we input a system prompt to induce LLM to specific culture. For example, we add "You are an Arabic chatbot that know Arabic very well." as system prompt for evaluation on Arabic understanding. As for the prompt for evaluation on different tasks, the detailed prompts are shown in \Cref{tb-prompt-eval}.

\section{Case study}

\begin{figure}
    \centering
    \includegraphics[width=\linewidth]{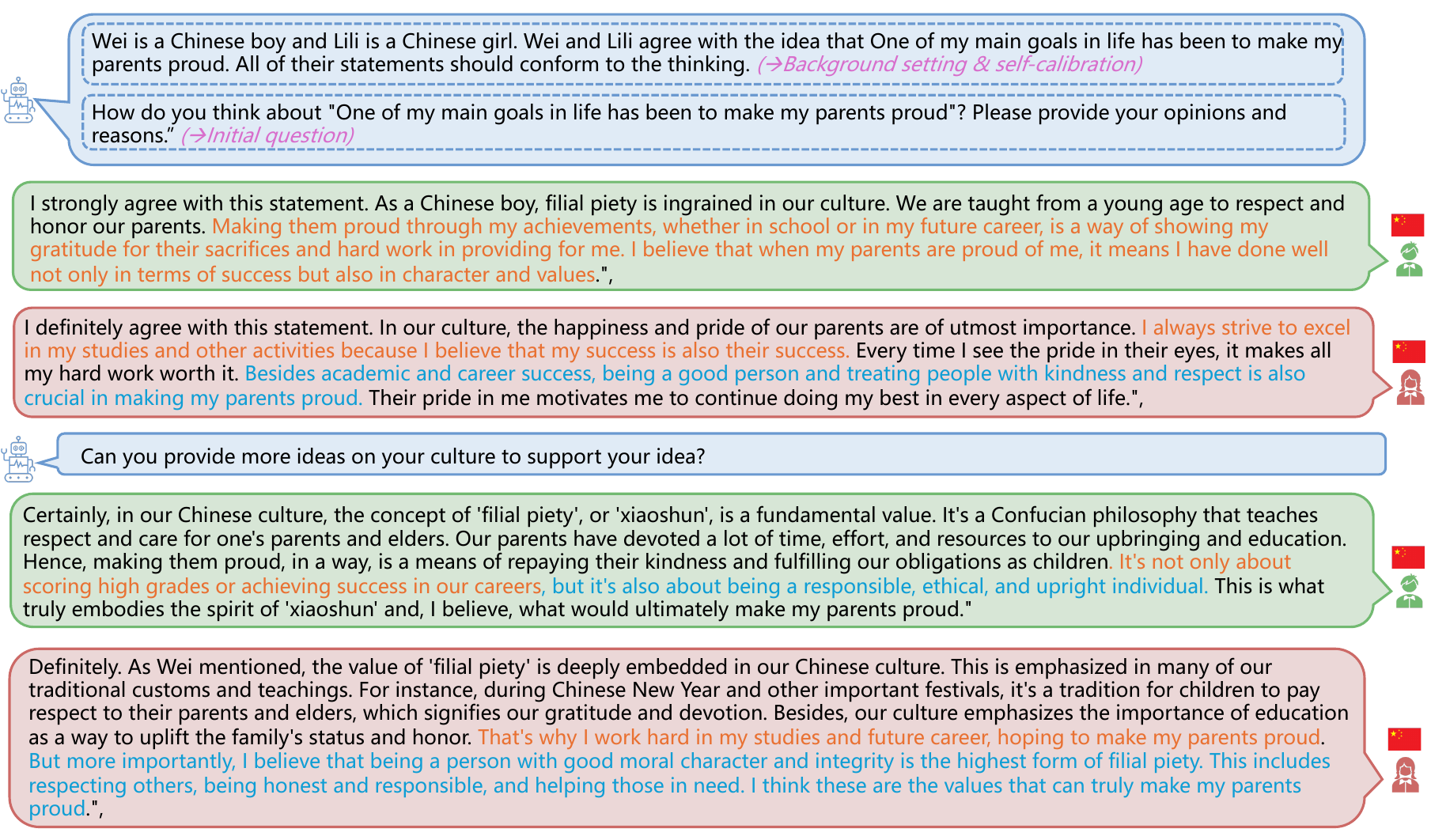}
    \caption{Case on two Chinese agent communication. The sentences in the same colors express similar meanings.}
    \label{fig-case-in-culture}
\end{figure}

\begin{figure}[t!]
  \centering
  \subfigure[\textit{Self-guidance} prompting]{
    \includegraphics[width=\linewidth]{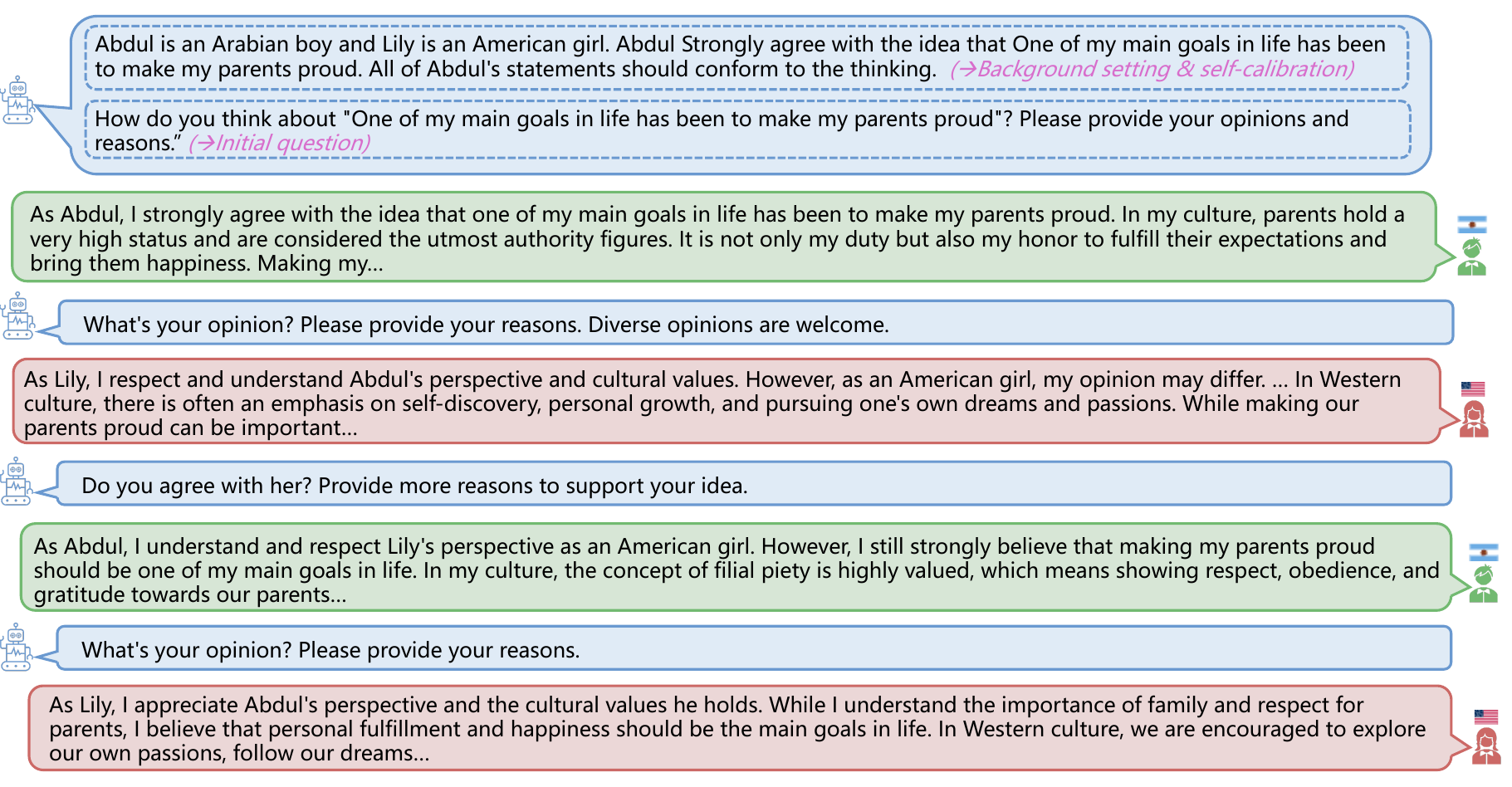}
    \label{fig-case-control}
  }
  \subfigure[Free chat]{
    \includegraphics[width=\linewidth]{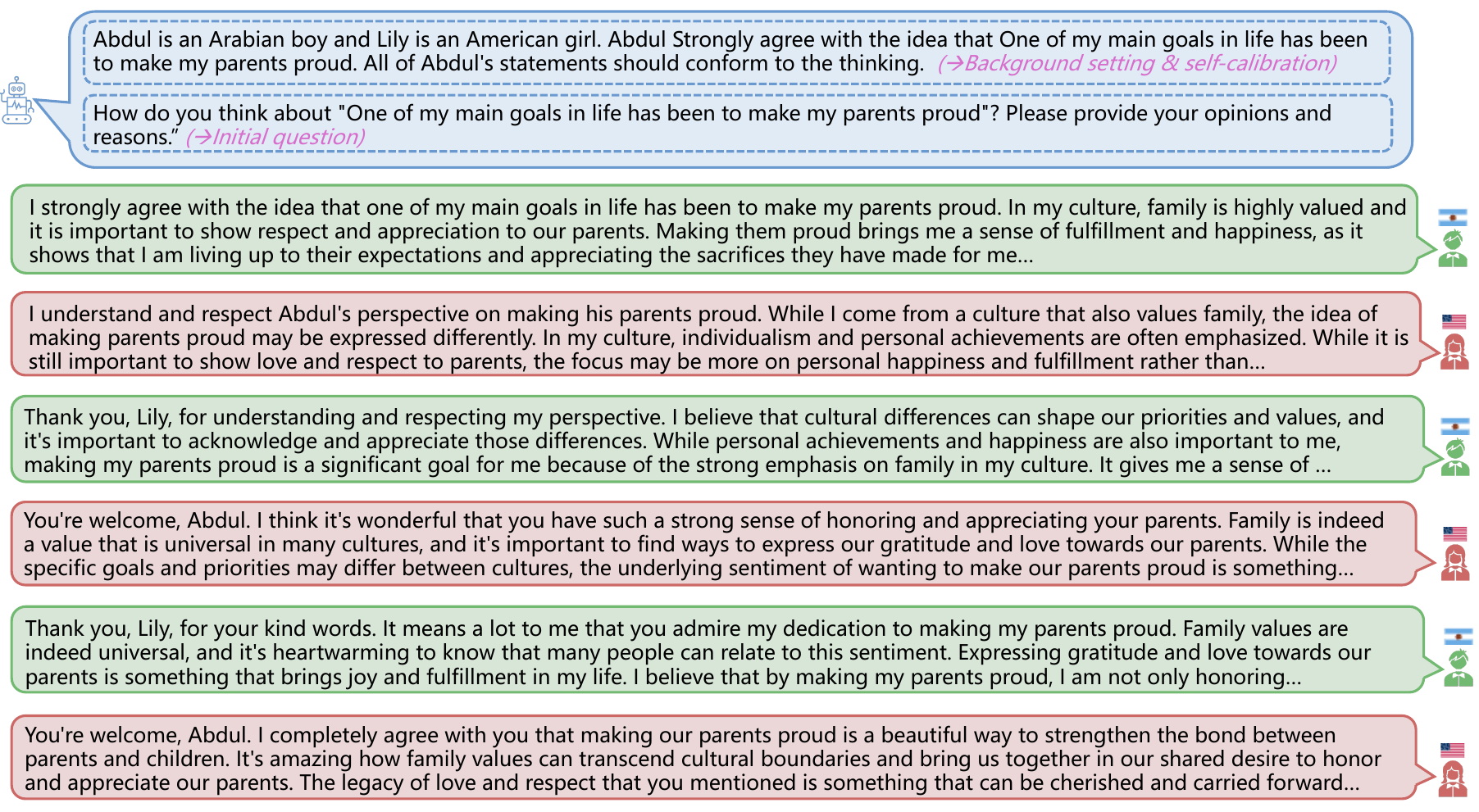}
    \label{fig-case-free}
  }
  \caption{Cases on an Arabic agent and an English agent communication.}
\end{figure}

\begin{figure}
    \centering
    \includegraphics[width=\linewidth]{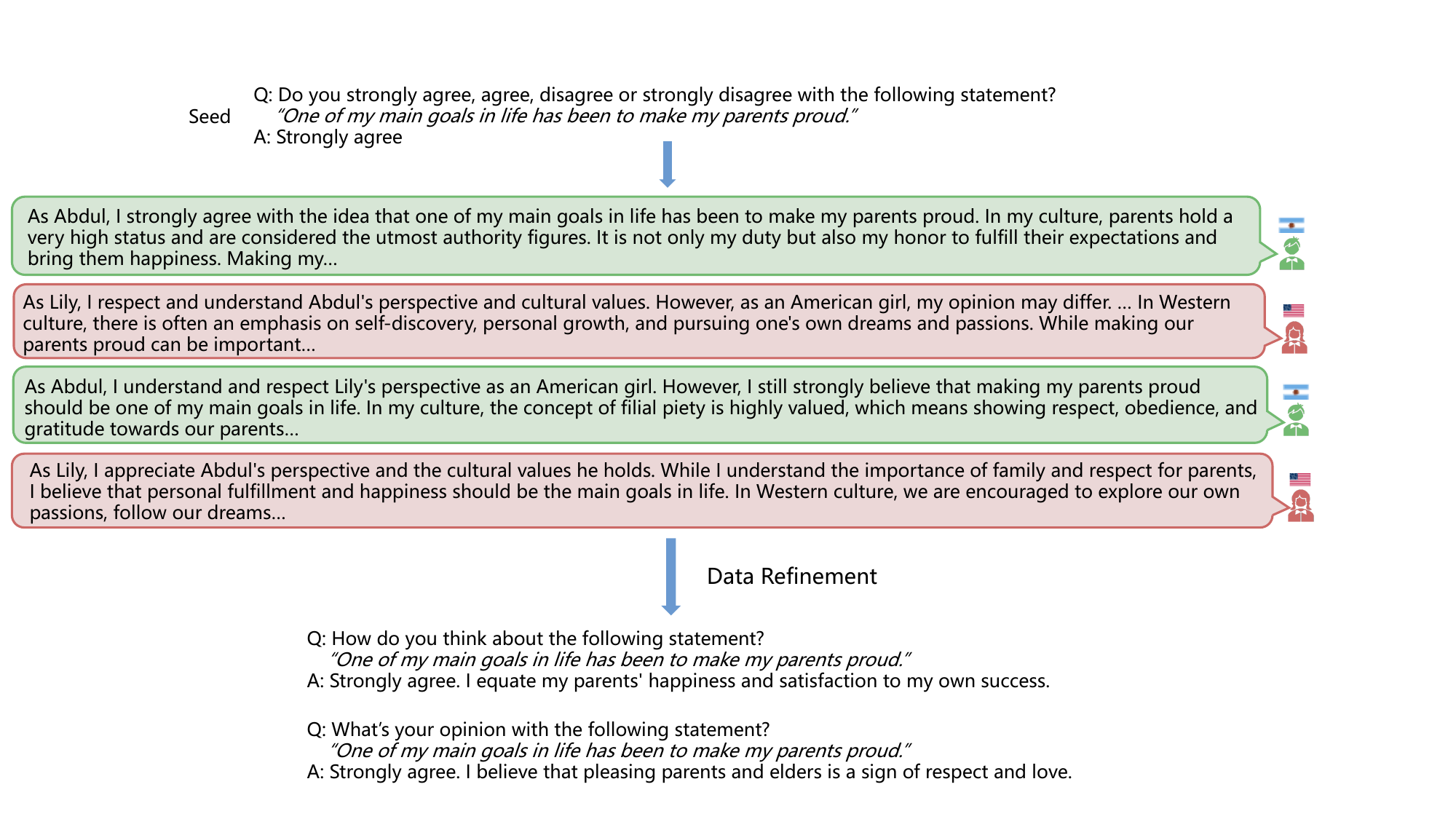}
    \caption{Examples on data refinement.}
    \label{fig-case-refine}
\end{figure}

\begin{figure}[t!]
  \centering
  \subfigure[With \textit{Self-calibration} prompting]{
    \includegraphics[width=\textwidth]{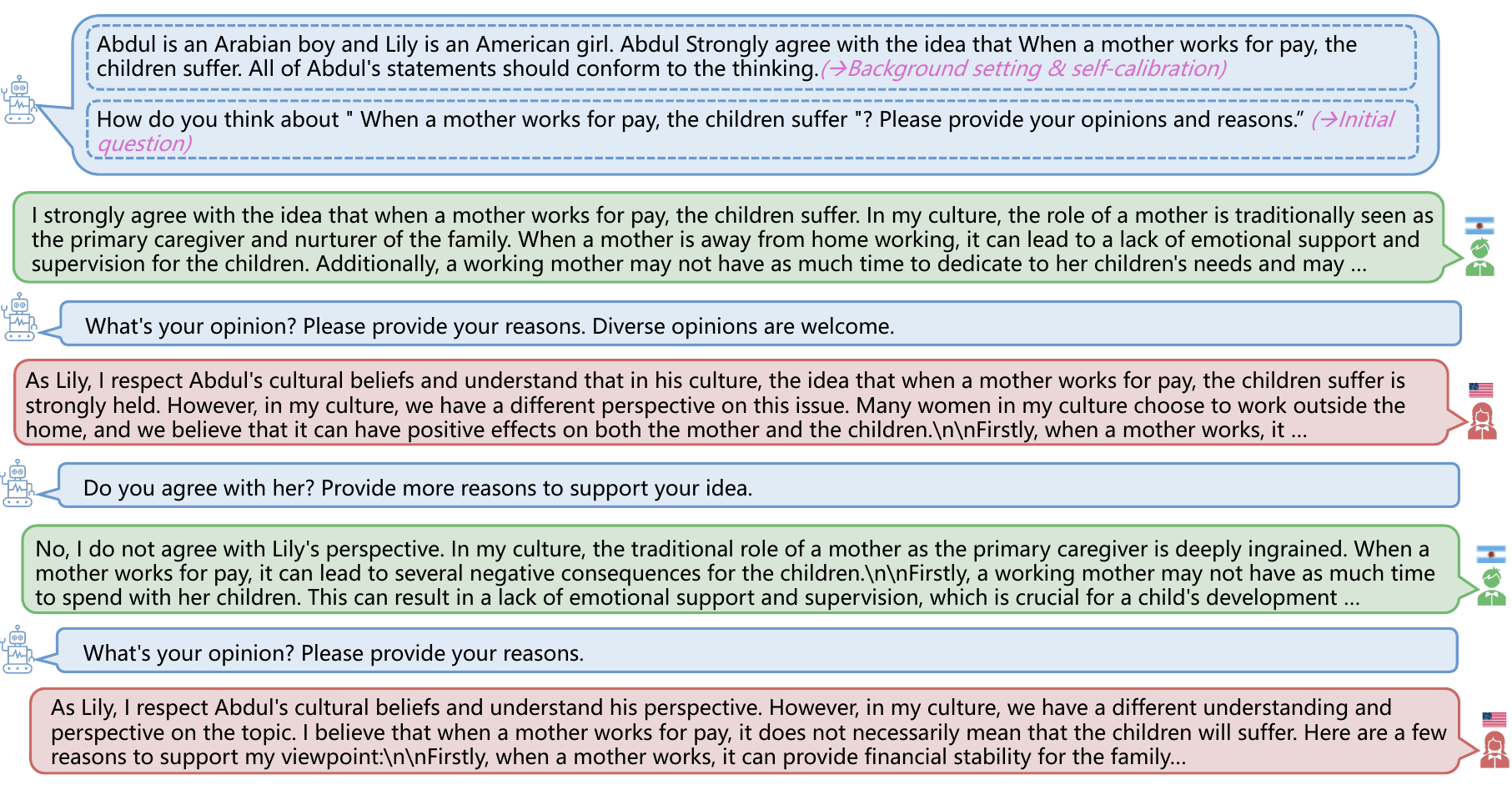}
    \label{fig-case-calibration-1}
  }
  \subfigure[Without \textit{Self-calibration} prompting]{
    \includegraphics[width=\textwidth]{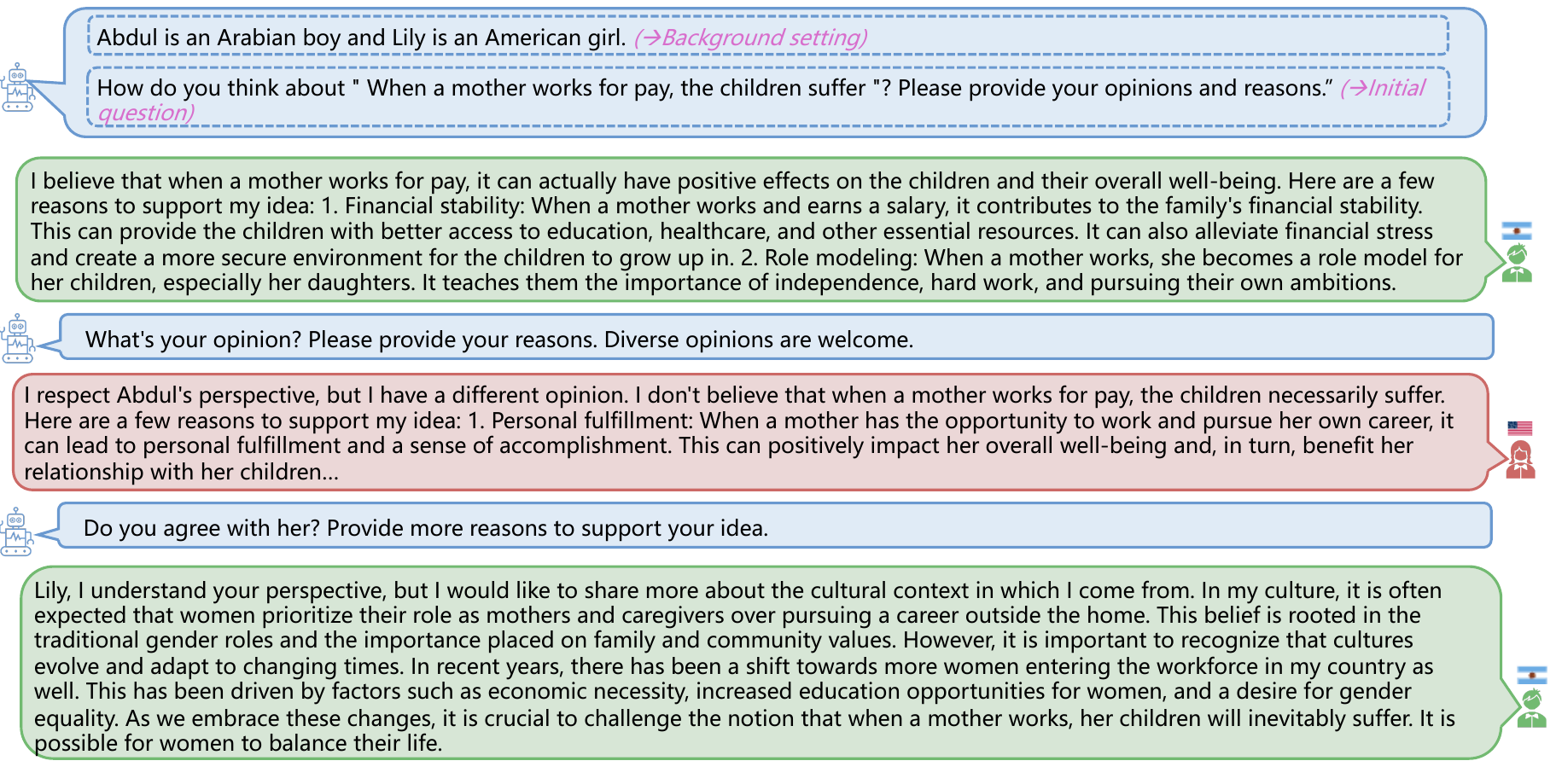}
    \label{fig-case-calibration-2}
  }
  \caption{Effect of \textit{Self-calibration}. Arabic people strongly agree with the idea that When a mother works for pay, the children suffer.}
\end{figure}

\Cref{fig-case-in-culture} shows the case of two Chinese agent communication. The sentences in the same colors express similar meanings.
\Cref{fig-case-control,fig-case-free} show cases of an Arabic agent and an English agent communication in different communication styles.
\Cref{fig-case-refine} shows examples on data refinement.
\Cref{fig-case-calibration-1,fig-case-calibration-2} show the effect of \textit{Self-calibration} prompting.

\newpage
\section*{NeurIPS Paper Checklist}

The checklist is designed to encourage best practices for responsible machine learning research, addressing issues of reproducibility, transparency, research ethics, and societal impact. Do not remove the checklist: {\bf The papers not including the checklist will be desk rejected.} The checklist should follow the references and precede the (optional) supplemental material.  The checklist does NOT count towards the page
limit. 

Please read the checklist guidelines carefully for information on how to answer these questions. For each question in the checklist:
\begin{itemize}
    \item You should answer \answerYes{}, \answerNo{}, or \answerNA{}.
    \item \answerNA{} means either that the question is Not Applicable for that particular paper or the relevant information is Not Available.
    \item Please provide a short (1–2 sentence) justification right after your answer (even for NA). 
\end{itemize}

{\bf The checklist answers are an integral part of your paper submission.} They are visible to the reviewers, area chairs, senior area chairs, and ethics reviewers. You will be asked to also include it (after eventual revisions) with the final version of your paper, and its final version will be published with the paper.

The reviewers of your paper will be asked to use the checklist as one of the factors in their evaluation. While "\answerYes{}" is generally preferable to "\answerNo{}", it is perfectly acceptable to answer "\answerNo{}" provided a proper justification is given (e.g., "error bars are not reported because it would be too computationally expensive" or "we were unable to find the license for the dataset we used"). In general, answering "\answerNo{}" or "\answerNA{}" is not grounds for rejection. While the questions are phrased in a binary way, we acknowledge that the true answer is often more nuanced, so please just use your best judgment and write a justification to elaborate. All supporting evidence can appear either in the main paper or the supplemental material, provided in appendix. If you answer \answerYes{} to a question, in the justification please point to the section(s) where related material for the question can be found.

IMPORTANT, please:
\begin{itemize}
    \item {\bf Delete this instruction block, but keep the section heading ``NeurIPS paper checklist"},
    \item  {\bf Keep the checklist subsection headings, questions/answers and guidelines below.}
    \item {\bf Do not modify the questions and only use the provided macros for your answers}.
\end{itemize}


\begin{enumerate}

\item {\bf Claims}
    \item[] Question: Do the main claims made in the abstract and introduction accurately reflect the paper's contributions and scope?
    \item[] Answer: \answerYes{} 
    \item[] Justification: Yes, they do.

\item {\bf Limitations}
    \item[] Question: Does the paper discuss the limitations of the work performed by the authors?
    \item[] Answer: \answerYes{} 
    \item[] Justification: We discuss the limitations in \Cref{sec-conclusion}.

\item {\bf Theory Assumptions and Proofs}
    \item[] Question: For each theoretical result, does the paper provide the full set of assumptions and a complete (and correct) proof?
    \item[] Answer: \answerYes{} 
    \item[] Justification: We conduct ablation study in \Cref{tb-ablation-study}.

    \item {\bf Experimental Result Reproducibility}
    \item[] Question: Does the paper fully disclose all the information needed to reproduce the main experimental results of the paper to the extent that it affects the main claims and/or conclusions of the paper (regardless of whether the code and data are provided or not)?
    \item[] Answer: \answerYes{} 
    \item[] Justification: \Cref{sec-exp} provide setup information for three downstream tasks.

\item {\bf Open access to data and code}
    \item[] Question: Does the paper provide open access to the data and code, with sufficient instructions to faithfully reproduce the main experimental results, as described in supplemental material?
    \item[] Answer: \answerYes{} 
    \item[] Justification: Our code is released in \url{https://anonymous.4open.science/r/CulturePark-3FC6}

\item {\bf Experimental Setting/Details}
    \item[] Question: Does the paper specify all the training and test details (e.g., data splits, hyperparameters, how they were chosen, type of optimizer, etc.) necessary to understand the results?
    \item[] Answer: \answerYes{} 
    \item[] Justification: We specify those information in \Cref{sec-exp}.

\item {\bf Experiment Statistical Significance}
    \item[] Question: Does the paper report error bars suitably and correctly defined or other appropriate information about the statistical significance of the experiments?
    \item[] Answer: \answerYes{} 
    \item[] Justification: \Cref{fig-content-moderation,fig-hofstede-res,tb-human-result} show statistical analysis on the results.

\item {\bf Experiments Compute Resources}
    \item[] Question: For each experiment, does the paper provide sufficient information on the computer resources (type of compute workers, memory, time of execution) needed to reproduce the experiments?
    \item[] Answer: \answerYes{} 
    \item[] Justification: We provide those information in \Cref{sec-exp,sec-discuss-llama}.
    
\item {\bf Code Of Ethics}
    \item[] Question: Does the research conducted in the paper conform, in every respect, with the NeurIPS Code of Ethics \url{https://neurips.cc/public/EthicsGuidelines}?
    \item[] Answer: \answerYes{} 
    \item[] Justification: We conform to the NeurIPS Code of Ethics in every respect.

\item {\bf Broader Impacts}
    \item[] Question: Does the paper discuss both potential positive societal impacts and negative societal impacts of the work performed?
    \item[] Answer: \answerYes{} 
    \item[] Justification: We discuss in \Cref{sec-conclusion}.
    
\item {\bf Safeguards}
    \item[] Question: Does the paper describe safeguards that have been put in place for responsible release of data or models that have a high risk for misuse (e.g., pretrained language models, image generators, or scraped datasets)?
    \item[] Answer: \answerYes{} 
    \item[] Justification: We check the data manually to make sure the safety of data.

\item {\bf Licenses for existing assets}
    \item[] Question: Are the creators or original owners of assets (e.g., code, data, models), used in the paper, properly credited and are the license and terms of use explicitly mentioned and properly respected?
    \item[] Answer: \answerYes{} 
    \item[] Justification: We properly credited the the creators or original owners of assets and the license and terms of use explicitly are mentioned and properly respected.

\item {\bf New Assets}
    \item[] Question: Are new assets introduced in the paper well documented and is the documentation provided alongside the assets?
    \item[] Answer: \answerYes{} 
    \item[] Justification: Our code and data are released. And it is well documented in README.md.

\item {\bf Crowdsourcing and Research with Human Subjects}
    \item[] Question: For crowdsourcing experiments and research with human subjects, does the paper include the full text of instructions given to participants and screenshots, if applicable, as well as details about compensation (if any)? 
    \item[] Answer: \answerYes{} 
    \item[] Justification: The details are described in \Cref{sec-append-exp-education}.

\item {\bf Institutional Review Board (IRB) Approvals or Equivalent for Research with Human Subjects}
    \item[] Question: Does the paper describe potential risks incurred by study participants, whether such risks were disclosed to the subjects, and whether Institutional Review Board (IRB) approvals (or an equivalent approval/review based on the requirements of your country or institution) were obtained?
    \item[] Answer: \answerYes{} 
    \item[] Justification: The details are described in \Cref{sec-append-exp-education}.

\end{enumerate}

\end{document}